\documentclass[11pt,a4paper]{article}
\usepackage[hyperref]{emnlp-ijcnlp-2019}
\usepackage{times}
\usepackage{latexsym}
\usepackage{booktabs}       % professional-quality tables
\usepackage{amsfonts}       % blackboard math symbols
\usepackage{nicefrac}       % compact symbols for 1/2, etc.
\usepackage{microtype}      % microtypography
\usepackage{amsmath}
\usepackage{todonotes}
\usepackage{xcolor}
\usepackage{tikz}
\usepackage{mathdots}
\usepackage{yhmath}
\usepackage{cancel}
\usepackage{color}
\usepackage{siunitx}
\usepackage{array}
\usepackage{multirow}
\usepackage{amssymb}
\usepackage{gensymb}
\usepackage{enumitem}
\usepackage{graphicx}
\usepackage{hyperref}
\usepackage{supertabular}
\usepackage{appendix}
\usepackage[caption=false]{subfig}
\newcommand{\xhdr}[1]{{\noindent\bfseries #1}.}

\newcommand{\shagun}[1]{\textcolor{purple}{Shagun: #1}}
\newcommand{\cut}[1]{}
\newcommand{\CITE}{\textcolor{red}{CITE}}

\usepackage{url}

\aclfinalcopy % Uncomment this line for the final submission

\title{CLUTRR: A Diagnostic Benchmark for Inductive Reasoning from Text}

\author{Koustuv Sinha \textsuperscript{1,3,4},
  Shagun Sodhani \textsuperscript{2,3},
  Jin Dong \textsuperscript{1,3}, \\ 
  {\bf Joelle Pineau \textsuperscript{1,3,4}} and
  {\bf William L. Hamilton \textsuperscript{1,3,4}} \\
  \textsuperscript{1} School of Computer Science, McGill University, Canada \\
  \textsuperscript{2} Universit\'e de Montr\'eal, Canada\\
  \textsuperscript{3} Montreal Institute of Learning Algorithms (Mila), Canada \\
  \textsuperscript{4} Facebook AI Research (FAIR), Montreal, Canada\\
  \{koustuv.sinha, sshagunsodhani, jin.dong, jpineau, wlh\}\\@\{mail.mcgill.ca, gmail.com, mail.mcgill.ca, cs.mcgill.ca, cs.mcgill.ca\}}

\date{}

\begin{document}
\maketitle
\begin{abstract}
  The recent success of natural language understanding (NLU) systems has been troubled by results highlighting the failure of these models to generalize in a systematic and robust way. 
  In this work, we introduce a diagnostic benchmark suite, named CLUTRR, to clarify some key issues related to the robustness and systematicity of NLU systems. 
  Motivated by classic work on inductive logic programming, CLUTRR requires that an NLU system infer kinship relations between characters in short stories.
  Successful performance on this task requires both extracting relationships between entities, as well as inferring the logical rules governing these relationships.
  CLUTRR allows us to precisely measure a model's ability for systematic generalization by evaluating on held-out combinations of logical rules, and it allows us to evaluate a model's robustness by adding curated noise facts.
  Our empirical results highlight a substantial performance gap between state-of-the-art NLU models (e.g., BERT and MAC) and a graph neural network model that works directly with symbolic inputs---with the graph-based model exhibiting both stronger generalization and greater robustness. 
\end{abstract}

\section{Introduction}
Natural language understanding (NLU) systems have been extremely successful at reading comprehension tasks, such as question answering (QA) and natural language inference (NLI). 
An array of existing datasets are available for these tasks. This includes datasets that test a system's ability to extract factual answers from text \citep{Rajpurkar2016-yc, Nguyen2016-ec, Trischler2016-fc, Mostafazadeh2016-hu, Su2016-so}, as well as datasets that emphasize commonsense inference, such as entailment between sentences \citep{bowman2015large, williams2018broad}. 

\begin{figure}[t]
\centering
\resizebox{0.45\textwidth}{!}{\includegraphics[]{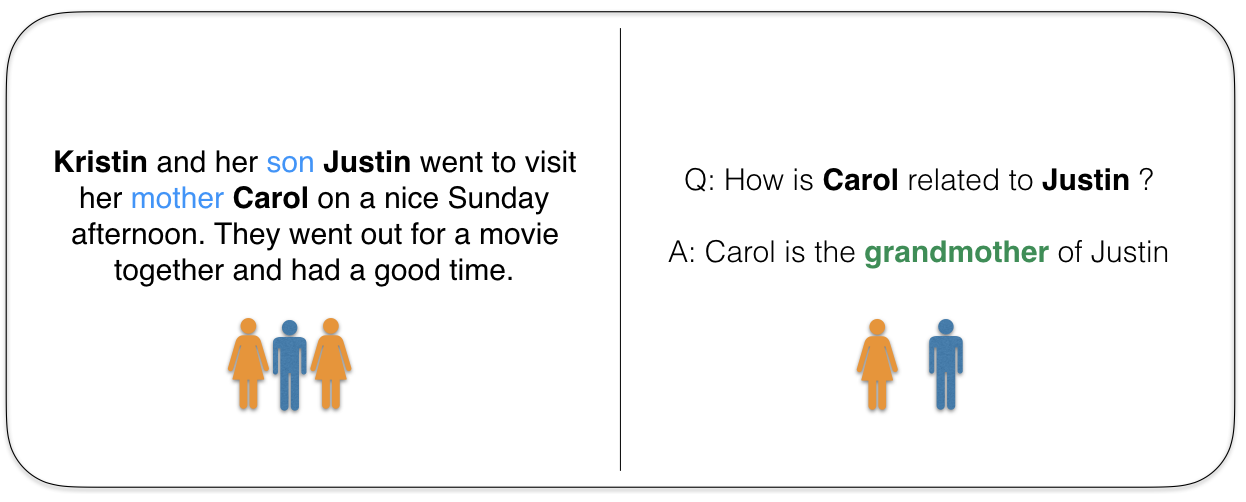}}
\caption{CLUTRR inductive reasoning task.}
\vspace{-15pt}
\label{fig:data_text}
\end{figure}

However, there are growing concerns regarding the ability of NLU systems---and neural networks more generally---to generalize in a systematic and robust way \cite{bahdanau2018systematic,lake2017generalization,Johnson2016-mw}.
For instance, recent work has highlighted the brittleness of NLU systems to adversarial examples \cite{jia2017adversarial}, as well as the fact that NLU models tend to exploit statistical artifacts in datasets, rather than exhibiting true reasoning and generalization capabilities \cite{gururangan2018annotation,kaushik2018much}.
These findings have also dovetailed with the recent dominance of large pre-trained language models, such as BERT, on NLU benchmarks \cite{devlin2018bert,peters2018deep}, which suggest that the primary difficulty in these datasets is incorporating the statistics of the natural language, rather than reasoning. 

An important challenge is thus to develop NLU benchmarks that can precisely test a model's capability for robust and systematic generalization.
Ideally, we want language understanding systems that can not only answer questions and draw inferences from text, but that can also do so in a systematic, logical, and robust way.   
While such reasoning capabilities are certainly required for many existing NLU tasks, most datasets combine several challenges of language understanding into one, such as co-reference/entity resolution, incorporating world knowledge, and semantic parsing---making it difficult to isolate and diagnose a model's capabilities for systematic generalization and robustness.

\xhdr{Our work}
Inspired by the classic AI challenge of inductive logic programming \cite{Quinlan1990-iv}---as well as the recently developed CLEVR dataset for visual reasoning \cite{Johnson2016-mw}---we propose a semi-synthetic benchmark designed to explicitly test an NLU model's ability for systematic and robust logical generalization. 

Our benchmark suite---termed CLUTRR (\textbf{C}ompositional \textbf{L}anguage \textbf{U}nderstanding and \textbf{T}ext-based \textbf{R}elational \textbf{R}easoning)---contains a large set of semi-synthetic stories involving hypothetical families.
Given a story, the goal is to infer the relationship between two family members, whose relationship is not explicitly mentioned (Figure \ref{fig:data_text}). 
To solve this task, a learning agent must extract the relationships mentioned in the text, induce the logical rules governing the kinship relationships (e.g., the transitivity of the sibling relation), and use these rules to infer the relationship between a given pair of entities.
Crucially, the CLUTRR benchmark allows us to test a learning agent's ability for {\em systematic generalization} by testing on stories that contain unseen combinations of logical rules.
CLUTRR also allows us to precisely test for the various forms of {\em model robustness} by adding different kinds of superfluous {\em noise facts} to the stories. 

We compare the performance of several state-of-the-art NLU systems on this task---including Relation Networks \cite{santoro2017simple}, Compositional Attention Networks \cite{hudson2018compositional} and BERT \cite{devlin2018bert}.
We find that the generalization ability of these NLU systems is substantially below that of a Graph Attention Network \cite{Velickovic2017-mh}, which is given direct access to symbolic representations of the stories. 
Moreover, we find that the robustness of the NLU systems generally does not improve by training on noisy data, whereas the GAT model is able to effectively learn robust reasoning strategies by training on noisy examples. 
Both of these results highlight important open challenges for closing the gap between machine reasoning models that work with unstructured text and models that are given access to more structured input.  

\section{Related Work}
\label{sec:related}
\cut{\shagun{Our work draws inspiration from and builds on work from various communities. Along the lines of classical work on inductive logic programming, we consider the task of inferring the kinship relations. Similar to reading comprehension benchmarks in NLP, we cast the underlying task as a question answering task given a kinship story. Keeping in pace with the recent research highlighting the challenge of systematic generalization in machine learning, we design the benchmark such that it can surface the relative strengths and shortcomings of the existing models and can also serve as a diagnostic tool.
}}

We draw inspiration from the classic work on inductive logic programming (ILP), a long line of reading comprehension benchmarks in NLP, as well as work combining language and knowledge graphs.

\xhdr{Reading comprehension benchmarks}
Many datasets have been proposed to test the reading comprehension ability of NLP systems. This includes the SQuAD \cite{Rajpurkar2016-yc}, NewsQA \cite{Trischler2016-fc}, and MCTest \cite{richardson2013mctest} benchmarks that focus on factual questions; the SNLI \cite{bowman2015large} and MultiNLI \cite{williams2018broad} benchmarks for sentence understanding; and the bABI tasks \cite{Weston2015-is}, to name a few. 
Our primary contribution to this line of work is the development of a carefully designed {\em diagnostic} benchmark to evaluate model robustness and systematic generalization in the context of NLU. 

\cut{
\xhdr{Systematic generalization}
A growing body of literature has demonstrated that NLU models tend to exploit statistical artifacts in datasets and lack true generalization capabilities \cite{jia2017adversarial,gururangan2018annotation, kaushik2018much, lake2017generalization}.
These critical examinations have dovetailed with similar studies on visual question answering \citep{agrawal2016analyzing,bahdanau2018systematic,Johnson2016-mw}.
CLUTRR, contributes to this growing area by introducing a principled and flexible benchmark to evaluate systematic generalization in the context of language understanding---with our notion of systematic generalization being grounded in classic work on inductive logic programming (ILP) \cite{Quinlan1990-iv}.
}

\xhdr{Question-answering with knowledge graphs}
Our work is also related to the domain of question answering and reasoning in knowledge graphs \citep{das2017go, xiong2018one, NIPS2018_7473, 8587330, xiong2017deeppath, welbl2018constructing, kartsaklis2018mapping}, where either the model is provided with a knowledge graph to perform inference over or where the model must infer a knowledge graph from the text itself. 
However, unlike previous benchmarks in this domain---which are generally {\em transductive} and focus on leveraging and extracting knowledge graphs as a source of background knowledge about a fixed set of entities---CLUTRR requires {\em inductive logical reasoning}, where every example requires reasoning over a new set of previously unseen entities.

\section{Benchmark Design}
\label{sec:benchmark_design}
In order to design an NLU benchmark that explicitly tests inductive reasoning and systematic generalization, we build upon the classic ILP task of inferring family (i.e., kinship) relations \cite{hinton1986learning, Muggleton1991-dh,Lavrac1994-pr,Kok2007-nb,Rocktaschel2017-ho}.
For example, given the facts that {\em ``Alice is Bob's mother''} and {\em ``Jim is Alice's father''}, one can infer with reasonable certainty that {\em ``Jim is Bob's grandfather.''}
While this example may appear trivial, it is a challenging task to design models that can learn from data to {\em induce} the logical rules necessary to make such inferences, %(e.g., the rule that the parent of a parent is generally a grandparent)
and it is even more challenging to design models that can systematically generalize by composing these induced rules.

Inspired by this classic task of logical induction and reasoning, the CLUTRR benchmark requires an NLU system to infer and reason about kinship relations by reading short stories.
Requiring that the models learn directly from natural language makes this task much more challenging than the purely symbolic ILP setting.
However, we leverage insights from traditional ILP to generate these stories in a semi-synthetic manner, providing precise control over the complexity of the reasoning required to solve the task. 

In its entirety, the CLUTRR benchmark suite allows researchers to generate diverse semi-synthetic short stories to test different aspects of inductive reasoning capabilities.
We publicly release the entire benchmark suite, including code to generate the semi-synthetic examples, the specific datasets that we introduce here, and the different baselines that we compare with.\footnote{Benchmark suite code can be obtained from \href{https://github.com/facebookresearch/clutrr}{https://github.com/facebookresearch/clutrr}. Generated datasets are available to view in \href{https://drive.google.com/file/d/1SEq_e1IVCDDzsBIBhoUQ5pOVH5kxRoZF/view?usp=sharing}{this link}.} 
%Sections \ref{subsec:data_gen} to \ref{subsec:queryrep} detail the data generating process that underlies the entire CLUTRR benchmark suite, while Section \ref{subsec:diag} describes the different datasets we experiment within this work. 

%\section{Benchmark Construction}
%\label{sec:dataset}

\begin{figure}[t]
\centering
\resizebox{0.5\textwidth}{!}{\includegraphics[]{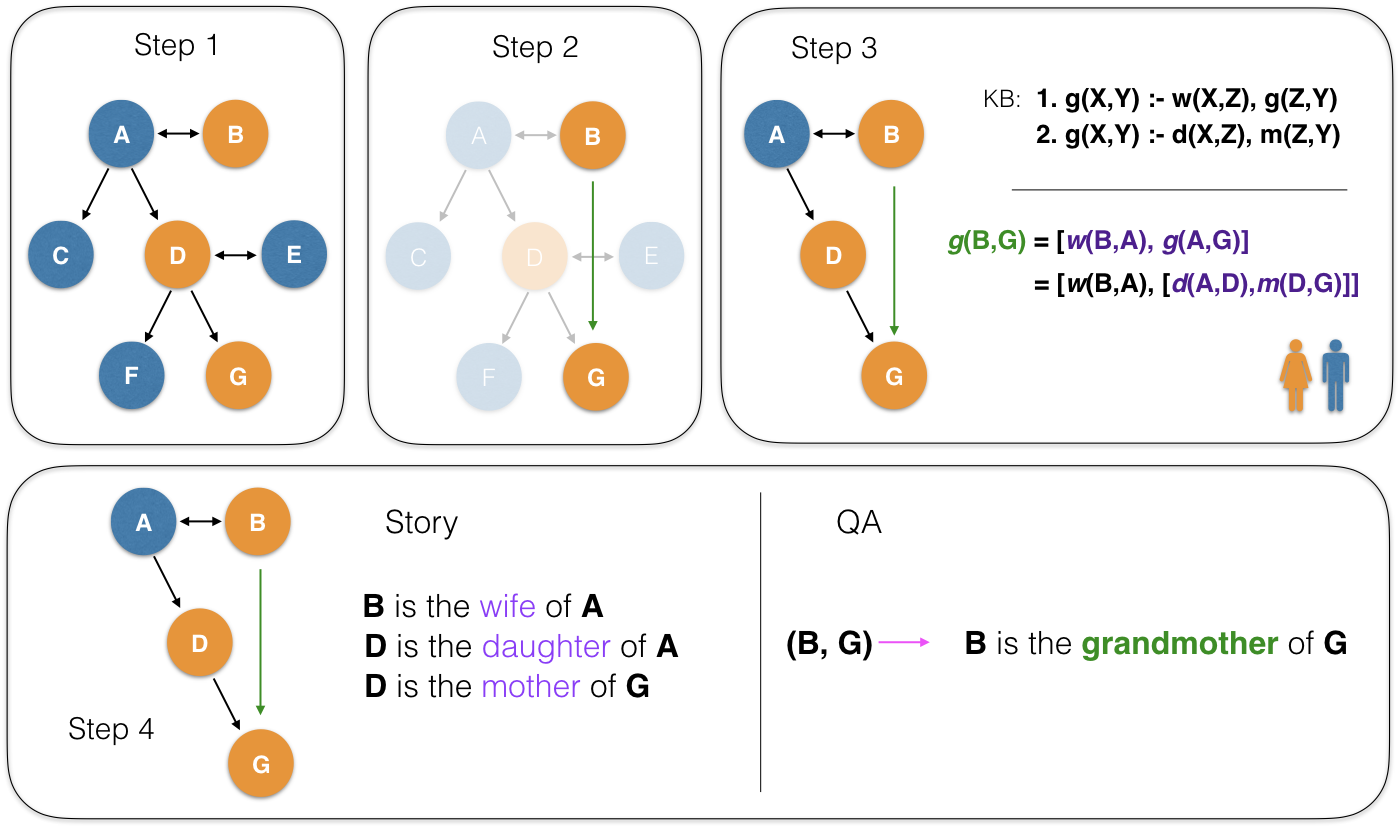}}
\caption{Data generation pipeline. Step 1: generate a kinship graph. Step 2: sample a target fact. Step 3: Use backward chaining to sample a set of facts. Step 4: Convert sampled facts to a natural language story.}
\vspace{-5pt}
\label{fig:data}
\end{figure}

\subsection{Overview of data generation process}\label{subsec:data_gen}
The core idea behind the CLUTRR benchmark suite is the following: Given a natural language story describing a set of kinship relations, the goal is to infer the relationship between two entities, whose relationship is {\em not} explicitly stated in the story.
To generate these stories, we first design a knowledge base (KB) with rules specifying how kinship relations resolve, and we use the following steps to create semi-synthetic stories based on this knowledge base:
\begin{enumerate}[topsep=3pt, parsep=6pt, leftmargin=34pt, itemsep=0pt, label={\bf Step \arabic*.}]
    \item \textbf{Generate} a random kinship graph that satisfies the rules in our KB.
    \item 
        \textbf{Sample a target fact} (i.e., relation) to predict from the kinship graph.
    \item
        \textbf{Apply backward chaining} to sample a set of facts that can prove the target relation (and optionally sample a set of ``distracting'' or ``irrelevant'' noise facts). 
    \item
        \textbf{Convert the sampled facts into a natural language story} through pre-specified text templates and crowd-sourced paraphrasing.
\end{enumerate}
Figure \ref{fig:data} provides a high-level overview of this idea, and the following subsections describe the data generation process in detail, as well as the diagnostic flexibility afforded by CLUTRR.

\subsection{Story generation}
\label{subsec:data_const}
The short stories in CLUTRR are essentially narrativized renderings of a set of logical facts.
In this section, we describe how we sample the logical facts that make up a story by generating random kinship graphs and using backward chaining to produce logical reasoning chains. 
The conversion from logical facts to natural language narratives is then described in Section \ref{subsec:nat_lang}.

\xhdr{Terminology and background}
Following standard practice in formal semantics, we use the term \textit{atom} to refer to a \textit{predicate} symbol and a list of terms, such as $[\texttt{grandfatherOf},X,Y]$, where the predicate $\texttt{grandfatherOf}$ denotes the \textit{relation} between the two \textit{variables}, $X$ and $Y$. We restrict the predicates to have an arity of 2, i.e.,  binary predicates. 
A logical \textit{rule} in this setting is of the form $\mathcal{H} \vdash \mathcal{B}$, where $\mathcal{B}$ is the \textit{body} of the rule, i.e., a conjunction of two \textit{atoms} ($[\alpha_1,\alpha_2]$) and $\mathcal{H}$ is the \textit{head}, i.e., a single \text{atom} ($\alpha$) that can be viewed as the goal or query. 
For instance, given a knowledge base (KB) $R$ that contains the single rule  $[\texttt{grandfatherOf},X,Y] \vdash [[\texttt{fatherOf},X,Z], [\texttt{fatherOf}, Z,Y]]$, the query $[\texttt{grandfatherOf},X,Y]$ evaluates to true if and only if the body $\mathcal{B}=[[\texttt{fatherOf},X,Z], [\texttt{fatherOf}, Z,Y]]$ is also true in a given world.
A rule is called a \textit{grounded} rule if all atoms in the rule are themselves \textit{grounded}, i.e., all variables are replaced with \textit{constants} or entities in a world. A \textit{fact} is a grounded binary predicate. A \textit{clause} is a conjunction of two or more atoms ($\mathcal{C}=(\mathcal{H}_{\mathcal{C}} \vdash \mathcal{B}_{\mathcal{C}} = ([\alpha_1,...,\alpha_n]))$) which can be built using a set of rules.

In the context of our data generation process, we distinguish between the knowledge base, $R$, which contains a finite number of predicates and rules specifying how kinship relations in a family resolve, and a particular kinship graph $G$, which contains a grounded set of atoms specifying the particular kinship relations that underlie a single story. 
In other words, $R$ contains the logical rules that govern all the generated stories in CLUTRR, while $G$ contains the grounded facts that underlie a specific story. 
%We restrict the set of rules in $R$ to consist of two atoms in the body (see the Appendix for the full set of rules). 
% Using the logical rules in $R$ (see the Appendix for the full set of rules), we build a \textit{clause} which can be a conjunction of one or multiple rules.  %Further, we consider only binary predicates that make it easy to define a graph structure on the story.

\xhdr{Graph generation}
To generate the kinship graph $G$ underlying a particular story, we first sample a set of gendered\footnote{Kinship and gender roles are oversimplified in our data (compared to the real world) to maintain tractability.} entities and kinship relations using a stochastic generation process. 
This generation process contains a number of tunable parameters---such as the maximum number of children at each node, the probability of an entity being married to another entity, etc.---and is designed to produce a valid, but possibly incomplete ``backbone graph''.
For instance, this backbone graph generation process will specify ``parent''/``child'' relations between entities but does not add ``grandparent'' relations. 
After this initial generation process, we recursively apply the logical rules in $R$ to the backbone graph to produce a final graph $G$ that contains the full set of kinship relations between all the entities.

\xhdr{Backward chaining}
The resulting graph $G$ provides the \textit{background knowledge} for a specific story, as each edge in this graph can be treated as a grounded predicate (i.e., fact) between two entities.
From this graph $G$, we sample the facts that make up the story, as well as the target fact that we seek to predict:
First, we (uniformly) sample a target relation $\mathcal{H}_{\mathcal{C}}$, which is the fact that we want to predict from the story.
Then, from this target relation $\mathcal{H}_{\mathcal{C}}$,  we run a simple variation of the backward chaining \citep{gallaire1978logic} algorithm for $k$ iterations starting from $\mathcal{H}_{\mathcal{C}}$, where at each iteration we uniformly sample a subgoal to resolve and then uniformly sample a KB rule that resolves this subgoal. 
Crucially, unlike traditional backward chaining, we do not stop the algorithm when a proof is obtained; instead, we run for a fixed number of iterations $k$ in order to sample a set of $k$ facts $\mathcal{B}_{\mathcal{C}}$ that imply the target relation $\mathcal{H}_{\mathcal{C}}$. 

\begin{figure}[t]
\centering
\resizebox{0.45\textwidth}{!}{\includegraphics[]{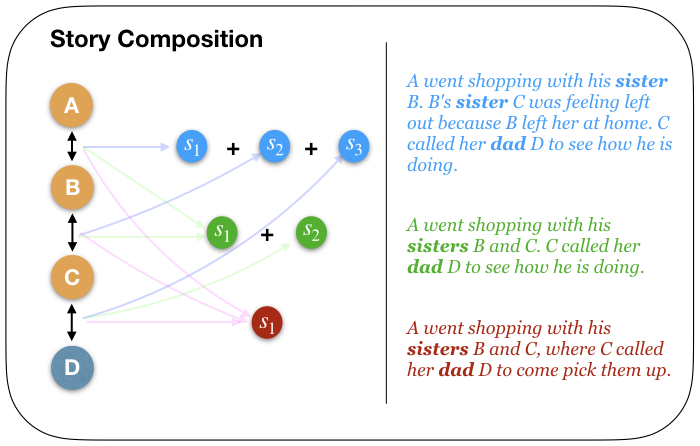}}
\caption{Illustration of how a set of facts can split and combined in various ways across sentences.}
\vspace{-10pt}
\label{fig:lang_composition}
\end{figure}

\subsection{Adding natural language}
\label{subsec:nat_lang}

So far, we have described the process of generating a  conjunctive logical clause $\mathcal{C}=(\mathcal{H}_{\mathcal{C}} \vdash \mathcal{B}_{\mathcal{C}})$, where $\mathcal{H}_{\mathcal{C}}=[\alpha^*]$ is the target fact (i.e., relation) we seek to predict and $\mathcal{B}_{\mathcal{C}} = [\alpha_1, ..., \alpha_k]$ is the set of supporting facts that imply the target relation.
We now describe how we convert this logical representation to natural language through crowd-sourcing. 

\xhdr{Paraphrasing using Amazon Mechanical Turk}
The basic idea behind our approach is that we show Amazon Mechanical Turk (AMT) crowd-workers the set of facts $\mathcal{B}_{\mathcal{C}}$ corresponding to a story and ask the workers to paraphrase these facts into a narrative. 
Since workers are given a set of facts $\mathcal{B}_{\mathcal{C}}$ to work from, they are able to combine and split multiple facts across separate sentences and construct diverse narratives (Figure \ref{fig:lang_composition}). 
Appendix 1.6 contains further details on our AMT interface (based on the ParlAI framework \cite{miller2017parlai}), data collection, and the quality controls we employed.

\xhdr{Reusability and composition}
One challenge for data collection via AMT is that the number of possible stories generated by CLUTRR grows combinatorially as the number of supporting facts increases, i.e., as  $k=|\mathcal{B}_\mathcal{C}|$ grows. 
This combinatorial explosion for large $k$---combined with the difficulty of maintaining the quality of the crowd-sourced paraphrasing for long stories---makes it infeasible to obtain a large number of paraphrased examples for $k>3$.
To circumvent this issue and increase the flexibility of our benchmark, we reuse and compose AMT paraphrases to generate longer stories. 
In particular, we collected paraphrases for stories containing $k=1,2,3$ supporting facts and then replaced the entities from these collected stories with placeholders in order to re-use them to generate longer semi-synthetic stories. 
An example of a story generated by stitching together two shorter paraphrases is provided below:
\vspace{-5pt}
\begin{quote}{\small
    [Frank] went to the park with his father, [Brett]. [Frank] called his brother [Boyd] on the phone. He wanted to go out for some beers. 
    [Boyd] went to the baseball game with his son [Jim].\\
    Q: What is [Brett] and [Jim]'s relationship?}
\end{quote}
\vspace{-5pt}
Thus, instead of simply collecting paraphrases for a fixed number of stories, we instead obtain a diverse collection of natural language templates that can be programmatically recombined to generate stories with various properties.

\begin{table}[]
\caption{Statistics of the AMT paraphrases. Jaccard word overlap is calculated within the templates of each individual clause of length $k$.}
\label{tab:placeholder}
\centering
\resizebox{0.48\textwidth}{!}{%
\begin{tabular}{@{}rrll@{}}
\toprule
\multicolumn{1}{l}{Number of Paraphrases} & \multicolumn{1}{l}{} &  & \# clauses \\ \midrule
\multicolumn{1}{r|}{} & $k$ = 1 & 1,868 & \multicolumn{1}{c}{20} \\
\multicolumn{1}{r|}{} & $k$ = 2 & 1,890 & \multicolumn{1}{c}{58} \\
\multicolumn{1}{r|}{} & $k$ = 3 & 2,258 & \multicolumn{1}{c}{236} \\ \cmidrule(l){2-4} 
\multicolumn{1}{r|}{} & Total & 6,016 &  \\ \midrule
\multicolumn{1}{l}{Unique Word Count} &  & 3,797 &  \\ \midrule
\multicolumn{1}{r|}{Jaccard Word Overlap} & Unigrams & 0.201 &  \\
\multicolumn{1}{r|}{} & Bigrams & 0.0385 &  \\ \bottomrule
\end{tabular}%
}
\end{table}

\xhdr{Dataset statistics}
At the time of submission, we have collected 6,016 unique paraphrases with an average of 19 paraphrases for every possible logical clause of length $k=1,2,3$. Table \ref{tab:placeholder} contains summary statistics of the collected paraphrases.
Overall, we found high linguistic diversity in the collected paraphrases.
For instance, the average Jaccard overlap in unigrams between pairs paraphrases corresponding to the same logical clause was only 0.201 and only 0.0385 for bigrams. 
The Appendix contains further examples of the paraphrases.

\xhdr{Human performance}
To get a sense of the data quality and difficulty involved in CLUTRR, we asked human annotators to solve the task for random examples of length $k=2,3,...,6$. 
We found that time-constrained AMT annotators performed well (i.e., ${>70\%}$) accuracy for ${k\leq 3}$ but struggled with examples involving longer stories, achieving 40-50\% accuracy for ${k > 3}$. However, trained annotators with unlimited time were able to solve 100\% of the examples (Appendix 1.7), highlighting the fact that this task requires attention and involved reasoning, even for humans.  

\subsection{Query representation and inference}\label{subsec:queryrep}

\xhdr{Representing the question}
The AMT paraphrasing approach described above allows us to convert the set of supporting facts $\mathcal{B}_\mathcal{C}$ to a natural language story, which can be used to predict the target relation/query $\mathcal{H}_\mathcal{C}$.
However, instead of converting the target query, $\mathcal{H}_\mathcal{C} = [\alpha^*]$, to a natural language question, we instead opt to represent the target query as a $K$-way classification task, where the two entities in the target relation are provided as input and the goal is to classify the relation that holds between these two entities. 
This representation avoids the pitfall of revealing information about the answer in the question \cite{kaushik2018much}.

\xhdr{Representing entities} 
When generating stories, entity names are randomly drawn from a set of 300 common gendered English names. 
Thus, depending on each run, the entities are never the same.
This ensures that the entity names are simply placeholders and uncorrelated from the task. 

\subsection{Variants of CLUTRR}\label{subsec:diag}
The modular nature of CLUTRR provides rich diagnostic capabilities for evaluating the robustness and generalization abilities of neural language understanding systems. 
We highlight some key diagnostic capabilities available via different variations of CLUTRR below.
These diagnostic variations correspond to the concrete datasets that we generated in this work, and we describe the results on these Datasets in Section \ref{sec:experiments}.

\xhdr{Systematic generalization} 
Most prominently, CLUTRR allows us to explicitly evaluate a model's ability for systematic generalization.
In particular, we rely on the following hold-out procedures to test systematic generalization:
\begin{itemize}[leftmargin=*, topsep=2pt, itemsep=2pt, parsep=2pt]
    \item During training, we hold out a subset of the collected paraphrases, and we only use this held-out subset of paraphrases when generating the test set. 
    Thus, to succeed on CLUTRR, an NLU system must exhibit {\em linguistic generalization} and be robust to linguistic variation at test time. 
    \item We also hold out a subset of the logical clauses during training (for clauses of length $k > 2$).\footnote{One should not holdout clauses from length $k=2$ in order to allow models to learn the compositionality of all possible binary predicates.}
    In other words, during training, the model sees all logical rules but does not see all {\em combinations} of these logical rules.
    Thus, in addition to linguistic generalization, success on this task also requires {\em logical generalization}.
    \item
    Lastly, as a more extreme form of both logical and linguistic generalization, we consider the setting where the models are trained on stories generated from clauses of length ${\leq k}$ and evaluated on stories generated from larger clauses of length ${>k}$. Thus, we explicitly test the ability for models to generalize on examples that require more steps of reasoning that any example they encountered during training.   
\end{itemize}
%We name the variation of the benchmark using these hold-out procedures \texttt{CLUTRR-Gen}.

\begin{figure}[t]
\centering
\resizebox{0.38\textwidth}{!}{\includegraphics[]{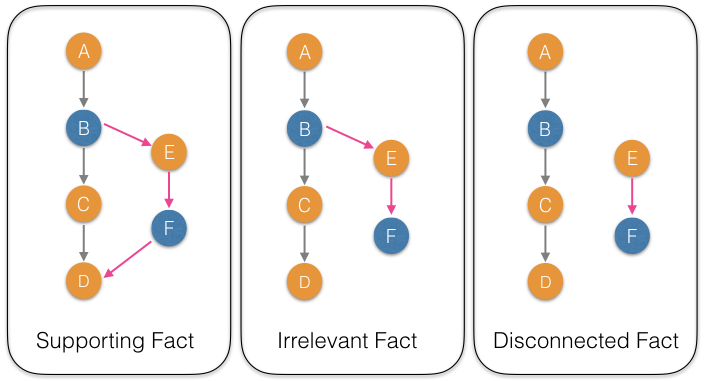}}
\vspace{-5pt}
\caption{Noise generation procedures of CLUTRR.}
\vspace{-12pt}
\label{fig:data_noise}
\end{figure}

\xhdr{Robust Reasoning}
In addition to evaluating systematic generalization, the modular setup of CLUTRR also allows us to diagnose model robustness by adding \textit{noise facts} to the generated narratives.
Due to the controlled semi-synthetic nature of CLUTRR, we are able to provide a precise taxonomy of the kinds of noise facts that can be added (Figure \ref{fig:data_noise}).
In order to structure this taxonomy, it is important to recall that any set of supporting facts $\mathcal{B}_\mathcal{C}$ generated by CLUTRR can be interpreted as a path, $p_\mathcal{C}$, in the corresponding kinship graph $G$ (Figure \ref{fig:data}).
Based on this interpretation, we view adding noise facts from the perspective of sampling three different types of noise paths, $p_n$, from the kinship graph $G$:
\begin{itemize}[  leftmargin=10pt, topsep=0pt, itemsep=0pt, parsep=0pt]
    \item \textit{Irrelevant facts}: We add a path $p_n$, which has exactly one shared end-point with $p_c$. In this way, this is a \textit{distractor} path, 
    which contains facts that are connected to one of the entities in the target relation, $\mathcal{H}_\mathcal{C}$, but do not provide any information that could be used to help answer the query. %We name this task \texttt{CLUTRR-Irrelevant} 
     \item \textit{Supporting facts}: 
    We add a path $p_n$, whose two end-points are on the path $p_\mathcal{C}$.
    The facts on this path $p_n$ are noise because they are not needed to answer the query, but they are supporting facts because they can, in principle, be used to construct alternative (longer) reasoning paths that connect the two target entities.
    %We denote this setup as \texttt{CLUTRR-Supporting}
    \item \textit{Disconnected facts}: We add paths which neither originate nor end in any entity on $p_c$. These disconnected facts involve entities and relations that are completely unrelated to the target query. 
    %We call this variant \texttt{CLUTRR-Disconnected}.
\end{itemize}

\section{Experiments}
\label{sec:experiments}
We evaluate several neural language understanding systems on the proposed CLUTRR benchmark to surface the relative strengths and shortcomings of these models in the context of inductive reasoning and combinatorial generalization.\footnote{Code to reproduce all the results in this section will be released at  \href{https://github.com/facebookresearch/clutrr/}{https://github.com/facebookresearch/clutrr/}.} We aim to answer the following key questions:
\begin{enumerate}[label=({\bf Q\arabic*}), leftmargin=28pt, topsep=0pt, itemsep=0pt, parsep=0pt]
\item How do state-of-the-art NLU models compare in terms of systematic generalization? Can these models generalize to stories with unseen combinations of logical rules?
 \item How does the performance of neural language understanding models compare to a graph neural network that has full access to graph structure underlying the stories? 
    \item How robust are these models to the addition of noise facts to a given story?
\end{enumerate}

\subsection{Baselines}
\label{sec:baselines}
\begin{figure*}[!htb]
     \centering
    \subfloat{{\includegraphics[width=0.48\textwidth]{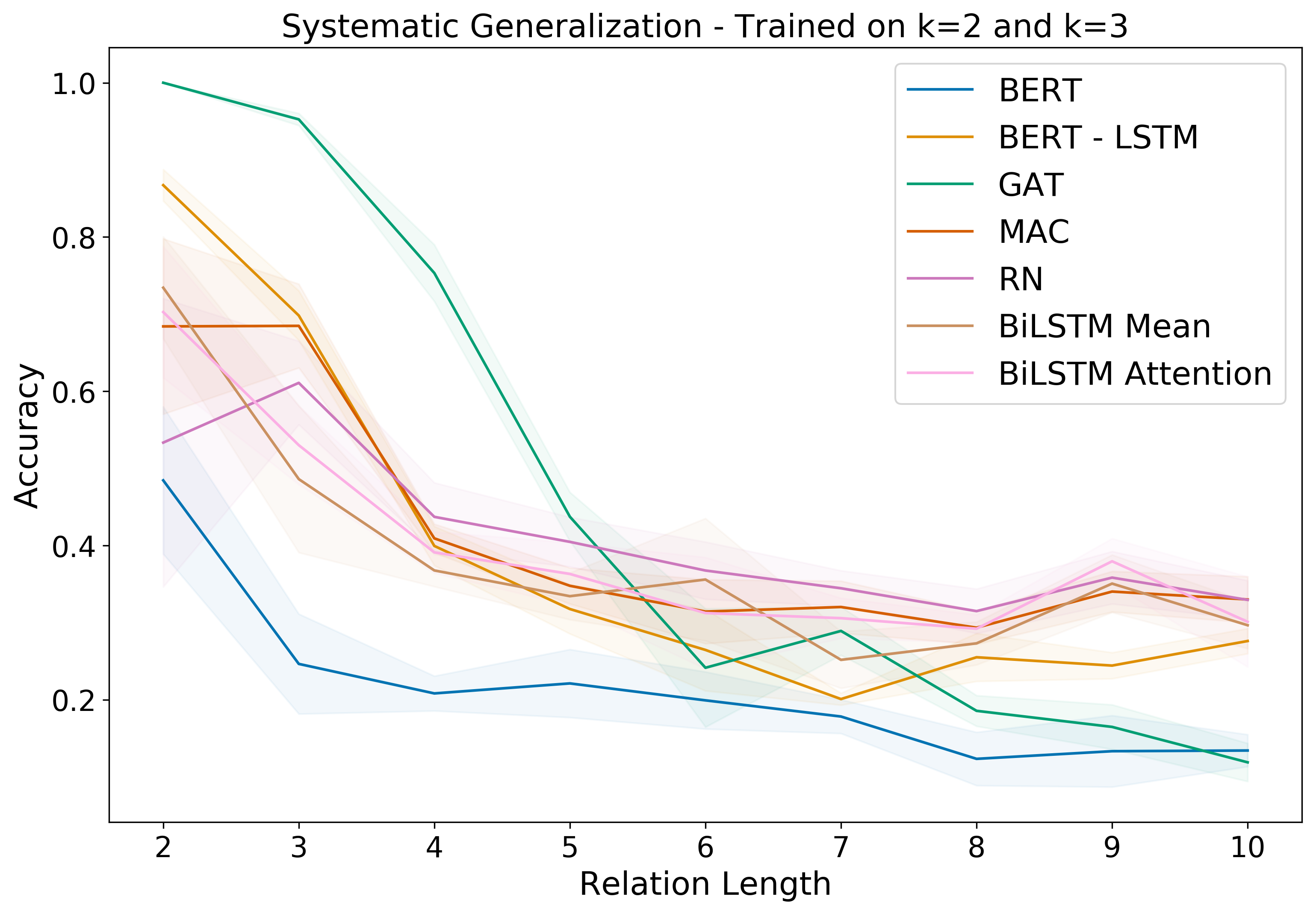} }}%
    \qquad
    \hspace{-20pt}
    \subfloat{{\includegraphics[width=0.48\textwidth]{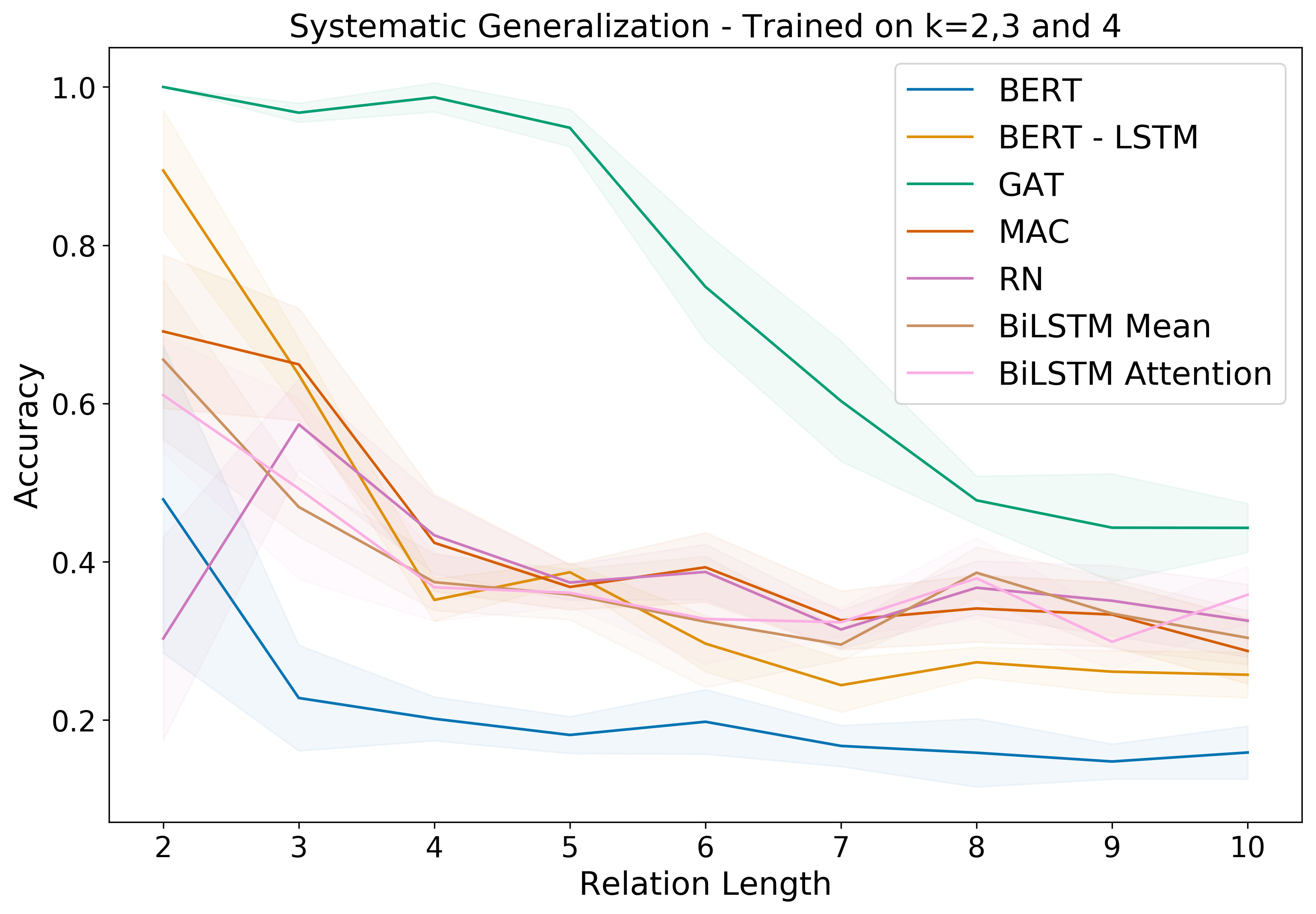} }}%
    \caption{Systematic generalization performance of different models when trained on clauses of length $k=2,3$ (Left) and $k=2,3,4$ (Right).}
    \vspace{-10pt}
    \label{fig:gen_1}
\end{figure*}

Our primary baselines are neural language understanding models that take unstructured text as input. 
We consider bidirectional LSTMs \citep{hochreiter1997long, cho2014learning} (with and without attention), as well as recently proposed models that aim to incorporate inductive biases towards relational reasoning: Relation Networks (RN) \citep{santoro2017simple}\cut{, Relational Recurrent Networks (RMC) \citep{santoro2018relational} } and Compositional Memory Attention Network (MAC) \citep{hudson2018compositional}. We also use the large pretrained language model, BERT \cite{devlin2018bert}, as well as a modified version of BERT having a trainable LSTM encoder on top of the pretrained BERT embeddings.
All of these models (except BERT) were re-implemented in PyTorch 1.0 \cite{paszke2017automatic} and adapted to work with the CLUTRR benchmark.

Since the underlying relations in the stories generated by CLUTRR inherently form a graph, we also experiment with a Graph Attention Network (GAT) \citep{Velickovic2017-mh}.
%Specifically, we consider }, as a representative model from the field of graph representation learning \cite{2017arXiv170905584H}. 
Rather than taking the textual stories as input, the GAT baseline receives a structured graph representation of the facts that underlie the story. 
%The motivation behind the inclusion of the GAT baseline is to evaluate the difficulty of the inductive reasoning task without the challenge of interpreting the natural language. 
%these two classes of models is to present the two extremes in current relational reasoning space : unstructured models which has to construct a latent graph, compared with structured model having the perfect graph as input.
%Future explorations can target the space between these models where an information retrieval (IR) model \citep{schoenmackers2010learning} can be used to extract the underlying graph from text and fed into a reasoning model.

% KS: I think the details of the models can go in the appendix as well

\xhdr{Entity and query representations}
We use the various baseline models to encode the natural language story (or graph) into a fixed-dimensional embedding.
With the exception of the BERT models, we do not use pre-trained word embeddings and learn the word embeddings from scratch using end-to-end backpropagation.
An important note, however, is that we perform Cloze-style anonymization \cite{hermann2015teaching} of the entities (i.e., names) in the stories, where each entity name is replaced by a \textit{@entity-k} placeholder, which is randomly sampled from a small, fixed pool of placeholder tokens. The embeddings for these placeholders are randomly initialized and fixed during training.\footnote{See Appendix 1.5 for a comparison of placeholder embedding approaches.}

To make a prediction about a target query given a story, we concatenate the embedding of the story (generated by the baseline model) with the embeddings of the two target entities and we feed this concatenated embedding to a 2-layer feed-forward neural network with a softmax prediction layer.

\subsection{Experimental Setup}
\label{sec:experimental_setup}

% In this work, using our benchmark suite we want to ask the following fundamental questions:

% \begin{itemize}
%     \item 
% \end{itemize}

\xhdr{Hyperparameters}
%All the baselines are implemented using PyTorch 1.0 \cite{paszke2017automatic}.
We selected hyperparameters for all models using an initial grid search on the systematic generalization task (described below).
All models were trained for 100 epochs with Adam optimizer and a learning rate of 0.001.
The Appendix provides details on the selected hyperparameters.

\xhdr{Generated datasets}
For all experiments, we generated datasets with 10-15k training examples.
In many experiments, we report training and testing results on stories with different clause lengths $k$.
(For brevity, we use the phrase ``clause length'' throughout this section to refer to the value $k=|\mathcal{B}_\mathcal{C}|$, i.e., the number of steps of reasoning that are required to predict the target query.)
In all cases, the training set contains 5000 train stories per $k$ value, and, during testing, all experiments use 100 test stories per $k$ value. 
All experiments were run 10 times with different randomly generated stories, and means and standard errors over these 10 runs are reported. 
As discussed in Section \ref{subsec:diag}, during training we holdout 20\% of the paraphrases, as well as 10\% of the possible logical clauses.

\subsection{Results and Discussion}
With our experimental setup in place, we now address the three key questions (\textbf{Q1-Q3}) outlined at the beginning of Section \ref{sec:experiments}.

\subsubsection*{Q1: Systematic Generalization}
\label{sec:sys_gen}

\begin{table*}[t!]
\caption{Testing the robustness of the various models when training and testing on stories containing various types of noise facts. The types of noise facts (supporting, irrelevant, and disconnected) are defined in Section \ref{subsec:diag}.}
\label{tab:robust}
\resizebox{\textwidth}{!}{
% Please add the following required packages to your document preamble:
% \usepackage{multirow}
\begin{tabular}{@{}cccccccc|c@{}}
\toprule
\multicolumn{1}{l}{} & \multicolumn{1}{l}{Models} & \multicolumn{6}{c}{Unstructured models (no graph)} & \multicolumn{1}{l}{Structured model (with graph)} \\ \midrule
\multicolumn{1}{l}{Training} & \multicolumn{1}{l|}{Testing} & BiLSTM - Attention & BiLSTM - Mean & RN & MAC & \multicolumn{1}{l}{BERT} & \multicolumn{1}{l|}{BERT-LSTM} & GAT \\ \midrule
Clean & \multicolumn{1}{c|}{Clean} & 0.58 \tiny $\pm 0.05$ & 0.53 \tiny $\pm 0.05$ & 0.49 \tiny $\pm 0.06$ & 0.63 \tiny $\pm 0.08$ & 0.37 \tiny $\pm 0.06$ & 0.67 \tiny $\pm 0.03$ & \textbf{1.0} \tiny $\pm 0.0$ \\ 
 & \multicolumn{1}{c|}{Supporting} & \textbf{0.76} \tiny $\pm 0.02$ & 0.64 \tiny $\pm 0.22$ & 0.58 \tiny $\pm 0.06$ & 0.71 \tiny $\pm 0.07$ & 0.28 \tiny $\pm 0.1$ & 0.66 \tiny $\pm 0.06$ & 0.24 \tiny $\pm 0.2$ \\
 & \multicolumn{1}{c|}{Irrelevant} & 0.7 \tiny $\pm 0.15$ & \textbf{0.76} \tiny $\pm 0.02$ & 0.59 \tiny $\pm 0.06$ & 0.69 \tiny $\pm 0.05$ & 0.24 \tiny $\pm 0.08$ & 0.55 \tiny $\pm 0.03$ & 0.51 \tiny $\pm 0.15$ \\
 & \multicolumn{1}{c|}{Disconnected} & 0.49 \tiny $\pm 0.05$ & 0.45 \tiny $\pm 0.05$ & 0.5 \tiny $\pm 0.06$ & 0.59 \tiny $\pm 0.05$ & 0.24 \tiny $\pm 0.08$ & 0.5 \tiny $\pm 0.06$ & \textbf{0.8} \tiny $\pm 0.17$ \\ \midrule
Supporting & \multicolumn{1}{c|}{Supporting} & 0.67 \tiny $\pm 0.06$ & 0.66 \tiny $\pm 0.07$ & 0.68 \tiny $\pm 0.05$ & 0.65 \tiny $\pm 0.04$ & 0.32 \tiny $\pm 0.09$ & 0.57 \tiny $\pm 0.04$ & \textbf{0.98} \tiny $\pm 0.01$ \\
\midrule
Irrelevant & \multicolumn{1}{c|}{Irrelevant} & 0.51 \tiny $\pm 0.06$ & 0.52 \tiny $\pm 0.06$ & 0.5 \tiny $\pm 0.04$ & 0.56 \tiny $\pm 0.04$ & 0.25 \tiny $\pm 0.06$ & 0.53 \tiny $\pm 0.06$ & \textbf{0.93} \tiny $\pm 0.01$ \\
\midrule
Disconnected & \multicolumn{1}{c|}{Disconnected} & 0.57 \tiny $\pm 0.07$ & 0.57 \tiny $\pm 0.06$ & 0.45 \tiny $\pm 0.11$ & 0.4 \tiny $\pm 0.1$ & 0.17 \tiny $\pm 0.05$ & 0.47 \tiny $\pm 0.06$ & \textbf{0.96} \tiny $\pm 0.01$ \\ \midrule
\multicolumn{1}{l}{Average} & \multicolumn{1}{c|}{} & \textbf{0.61} \tiny $\pm 0.08$ & 0.59 \tiny $\pm 0.08$ & 0.54 \tiny $\pm 0.07$ & \textbf{0.61} \tiny $\pm 0.06$ & 0.30 \tiny $\pm 0.07$ & 0.56 \tiny $\pm 0.05$ & \textbf{0.77} \tiny $\pm 0.09$ \\ \bottomrule
\end{tabular}}
\vspace{-5pt}
\end{table*}

We begin by using CLUTRR to evaluate the ability of the baseline models to perform systematic generalization (\textbf{Q1}).
In this setting, we consider two training regimes: in the first regime, we train all models with clauses of length $k=2,3$, and in the second regime, we train with clauses of length $k=2,3,4$.
We then test the generalization of these models on test clauses of length $k=2,...,10$. 

Figure \ref{fig:gen_1} illustrates the performance of different models on this generalization task.
We observe that the GAT model is able to perform near-perfectly on the held-out logical clauses of length $k=3$, with the BERT-LSTM being the top-performer among the text-based models but still significantly below the GAT. 
Not surprisingly, the performance of all models degrades monotonically as we increase the length of the test clauses, which highlights the challenge of ``zero-shot'' systematic generalization \cite{lake2017generalization, 2018arXiv181107017S}. 
However, as expected, all models improve on their generalization performance when trained on $k=2,3,4$ rather than just $k=2,3$ (Figure \ref{fig:gen_1}, right). The GAT, in particular, achieves the biggest gain by this expanded training.

\subsubsection*{Q2: The Benefit of Structure}
The empirical results on systematic generalization also provide insight into how the text-based NLU systems compare against the graph-based GAT model that has full access to the logical graph structure underlying the stories (\textbf{Q2}).
Indeed, the relatively strong performance of the GAT model (Figure \ref{fig:gen_1}) suggests that the language-based models fail to learn a robust mapping from the natural language narratives to the underlying logical facts. 

To further confirm this trend, we ran experiments with modified train and test splits for the text-based models, where the same set of natural language paraphrases were used to construct the narratives in both the train and test splits (see Appendix 1.3 for details). In this simplified setting, the text-based models must still learn to reason about held-out logical patterns, but the difficulty of parsing the natural language is essentially removed, as the same natural language paraphrases are used during testing and training. We found that the text-based models were competitive with the GAT model in this simplified setting (Appendix Figure 1), confirming that the poor performance of the text-based models on the main task is driven by the difficulty of parsing the unseen natural language narratives.

\label{sec:amt_par}

\subsubsection*{Q3: Robust Reasoning}
\label{sec:rob_reason}

Finally, we use CLUTRR to systematically evaluate how various baseline neural language understanding systems cope with noise (\textbf{Q3}). 
In all the experiments we provide a combination of $k=2$ and $k=3$ length clauses in training and testing, with noise facts being added to the train and/or test set depending on the setting (Table \ref{tab:robust}). 
We use the different types of noise facts defined in Section \ref{subsec:diag}.

Overall, we find that the GAT baseline outperforms the unstructured text-based models across most testing scenarios (Table \ref{tab:robust}), which showcases the benefit of a structured feature space for robust reasoning.
When training on clean data and testing on noisy data, we observe two interesting trends that highlight the benefits and shortcomings of the various model classes:
\begin{enumerate}[leftmargin=*, topsep=2pt, itemsep=0pt]
    \item All the text-based models excluding BERT actually perform better when testing on examples that have {\em supporting} or {\em irrelevant} facts added. This suggests that these models actually benefit from having more content related to the entities in the story. Even though this content is not strictly useful or needed for the reasoning task, it may provide some linguistic cues (e.g., about entity genders) that the models exploit. In contrast, the BERT-based models do not benefit from the inclusion of this extra content, which is perhaps due to the fact that they are already built upon a strong language model (e.g., that already adequately captures entity genders.)
    \item  The GAT model performs poorly when {\em supporting} facts are added but has no performance drop when {\em disconnected} facts are added. This suggests that the GAT model is sensitive to changes that introduce cycles in the underlying graph structure but is robust to the addition of noise that is disconnected from the target entities.
\end{enumerate}
Moreover, when we trained on noisy examples, we found that only the GAT model was able to consistently improve its performance (Table \ref{tab:robust}).
Again, this highlights the performance gap between the unstructured text-based models and the GAT.

% TODO: write about cycles
% Point raised by Shagun: cycles may be easily understood by Graph-background reviewers, but how best to put it so that general NLP reviewers understand the difference.

%We found that the MAC model was the overall strongest performer among the text-based models.
%However, the performance gap between 
%As a future work we hope to study human performance on our benchmark to effectively compare the difficulty in understanding systematicity and robust inductive reasoning in natural language.

% removed the section in appendix
 
%This provides yet another empirical proof that regularization---in this case provided via a form of data augemntation--- a model might be a key ingredient to generalization \cite{verma2018manifold}.

\section{Conclusion}
In this paper we introduced the CLUTRR benchmark suite to test the systematic generalization and inductive reasoning capababilities of NLU systems. 
We demonstrated the diagnostic capabilities of CLUTRR and found that existing NLU systems exhibit relatively poor robustness and systematic generalization capabilities---especially when compared to a graph neural network that works directly with symbolic input. 
These results highlight the gap that remains between machine reasoning models that work with unstructured text and models that are given access to more structured input. 
We hope that by using this benchmark suite, progress can be made in building more compositional, modular, and robust NLU systems. 
%By rigorous experiments and comparing several model performance we show the utility of this benchmark, which helps us to critically examine the issues pertaining to language understanding and systematicity. 
%We hope that by using this benchmark progress can be made in building more compositional and modular systems, which can learn the abstract rules governing a task without overfitting on a given dataset.
\section{Acknowledgements}
The authors would like to thank Jack Urbanek, Stephen Roller, Adina Williams, Dzmitry Bahdanau, Prasanna Parthasarathy, Harsh Satija for useful discussions and technical help. The authors would also like to thank Abhishek Das, Carlos Eduardo Lassance, Gunshi Gupta, Milan Aggarwal, Rim Assouel, Weiping Song, and Yue Dong for feedback on the draft. The authors also like to thank the many anonymous Mechanical Turk participants for providing paraphrases, and thank Sumana Basu, Etienne Denis, Jonathan Lebensold, and Komal Teru for providing human performance measures. The authors would also like to thank Sanghyun Yoo, Jehun Jeon and Dr Young Sang Choi of Samsung Advanced Institute of Technology (SAIT) for supporting the previous workshop version of this work. The authors are grateful to Facebook AI Research (FAIR) for providing extensive compute and GPU resources and support.
This research was supported by the Canada CIFAR Chairs in AI program.

\bibliography{emnlp-ijcnlp-2019}
\bibliographystyle{acl_natbib}

\clearpage
\appendix
\renewcommand\thesection{\arabic{section}}
%\appendixpage

\section{Appendix}
\label{sec:appendix}
\subsection{Implementation details of the baseline models}
\label{app:baselines}

\xhdr{Setup} We implemented all the models using the encoder-decoder architecture. The encoders are different baseline models (listed below).The encoder takes as input the given story (paragraph ) $p=(p_1, p_2, ...)$ and produces the representation of the story. In all the models, the decoder is implemented as a 2-layer MLP which takes as input the concatenated representation of the story and the embedding of the entities (for which the relationship is to be predicted) and returns a softmax distribution over the relation types. We now describe the different baseline models (encoders) in detail:

\noindent \textbf{LSTM} \cite{hochreiter1997long}: The input paragraph is processed by a two-layer Bidirectional LSTM and the hidden state corresponding to the last time-step is used as the representation of the story.

\noindent\textbf{LSTM+attention} \cite{cho2014learning}: Similar to LSTM, but instead of using just the hidden state at the last timestep, the model computes the attention-weighted mean of the hidden state at all time steps to use as the representation of the story.

\noindent\textbf{Relation Networks - RN} \cite{santoro2017simple}: An relation module (implemented as an MLP) is used alongside the LSTM to learn   pairwise relations among all the pairs of sentences. These relation representations are the output of the relational module. Our input data is prepared as a batch of $sentences \times words$. Each sentence is fed to the LSTM, followed by a pooling (e.g. mean, max) over all hidden states of each sentence to generate the sentence embeddings. The query embeddings are no longer needed in the decoder since they have  been incorporated by the relational module when learning  relations between sentences.

\noindent\textbf{MAC}(Compositional Attention Network) \cite{hudson2018compositional}: A MAC cell is similar to RN, but it also which contains a \textit{control} state $c$ and \textit{memory} state $m$ and can iterate over the input several times. The number of iterations is a hyperparameter. Just like RN, MAC is added behind the LSTM. In each iteration, the model attends to the embeddings of the query entities to generate the current control $c_i$. Another attention head over $c_i$ and all hidden outputs of LSTM is used to distill the new information $r_i$. In the end, a linear layer is used to generate the new memory $m_i$ by combining $r_i$ and $m_{i-1}$. The final memory state gives the representation of the story. This model is the state-of-the-art model for the CLEVER task.

% but the relational bias is incorporated by attention-based iterations over both query and paragraph. State-of-the-art on the CLEVER task.

\noindent \textbf{BERT} \cite{devlin2018bert}: We adapt BERT pretrained language model to our task. Specifically, we use two variants of BERT - the vanilla 12-layered frozen BERT with pre-trained embeddings, and BERT-LSTM, where a one-layer LSTM encoder is added on top of pretrained BERT embeddings. BERT encodes the sentences into 768-dimensional vectors. To ensure that BERT does not treat the entities as unknown tokens (and hence producing the same representation for all of them), we represent the entities with numbers in the vanilla BERT setup. In BERT-LSTM, we replace the entity embeddings by our entity embedding lookup policy (Refer Appendix \ref{app:embeddings}). In both the cases, we use a simple two-layer MLP decoder which takes as inputs the pooled document representation and query representations and produces the softmax distribution over the relations.

\noindent\textbf{Graph Attention Network(GAT)} \cite{Velickovic2017-mh}: Entity(modelled as nodes in the graph ) representations are learned by using the GAT Graph Neural Network with attention-based aggregation over the neighbor nodes. We modify the GAT architecture by attending over each node $v_j$ in the neighborhood of $v_i$ by concatenating the edge representation $e_{i,j}$ to the representation of $v_i$.

\noindent\textbf{Relational Recurrent Network (RMC)} \cite{santoro2018relational}: We also implemented RMC, a recently proposed model for relational reasoning. It works like an RNN, processing words step-by-step, except that a memory matrix ($num\_slots \times mem\_size$) is added as the hidden state.  The relational bias is extracted in each step by using self-attention over the concatenation of memory matrix and the word input in the current step. The final memory matrix is the representation of the story. Our implementation is based on another open source implementation. \footnote{\href{https://github.com/L0SG/relational-rnn-pytorch}{https://github.com/L0SG/relational-rnn-pytorch}}. We noticed that the performance of the model is significantly less across all the tasks by a large margin. Till the time of the submission, we could not verify whether this subpar performance is due to buggy implementation of the code or due to some unexplored hyperparameter combination. Hence we decided not to include the results corresponding to this model in the empirical evaluation. We will continue working on verifying the implementation of the model.

\subsection{Relations and KB used in CLUTRR Benchmark}
\label{app:kb_rules}

\small
\begin{align*}
\begin{split}
    [\texttt{grand}, X,Y] &\vdash [[\texttt{child}, X,Z],[\texttt{child}, Z,Y]], \\
    [\texttt{grand}, X,Y] &\vdash [[\texttt{SO}, X,Z],[\texttt{grand}, Z,Y]], \\
    [\texttt{grand}, X,Y] &\vdash [[\texttt{grand}, X,Z], \\ & [\texttt{sibling}, Z,Y]], \\
    [\texttt{inv-grand}, X,Y] &\vdash [[\texttt{inv-child}, X,Z], \\ & [\texttt{inv-child}, Z,Y]], \\
    [\texttt{inv-grand}, X,Y] &\vdash [[\texttt{sibling}, X,Z], \\ & [\texttt{inv-grand}, Z,Y]], \\
    [\texttt{child}, X,Y] &\vdash [[\texttt{child}, X,Z],\\ & [\texttt{sibling}, Z,Y]], \\
    [\texttt{child}, X,Y] &\vdash [[\texttt{SO}, X,Z],\\ & [\texttt{child}, Z,Y]], \\
    [\texttt{inv-child}, X,Y] &\vdash [[\texttt{sibling}, X,Z], \\ & [\texttt{inv-child}, Z,Y]], \\
    [\texttt{inv-child}, X,Y] &\vdash [[\texttt{child}, X,Z], \\ & [\texttt{inv-grand}, Z,Y]], \\
    [\texttt{sibling}, X,Y] &\vdash [[\texttt{child}, X,Z], \\ & [\texttt{inv-un}, Z,Y]], \\
    [\texttt{sibling}, X,Y] &\vdash [[\texttt{inv-child}, X,Z], \\ & [\texttt{child}, Z,Y]], \\
    [\texttt{sibling}, X,Y] &\vdash [[\texttt{sibling}, X,Z], \\ &[\texttt{sibling}, Z,Y]], \\
    [\texttt{in-law}, X,Y] &\vdash [[\texttt{child}, X,Z], \\ & [\texttt{SO}, Z,Y]], \\
    [\texttt{inv-in-law}, X,Y] &\vdash [[\texttt{SO}, X,Z], \\ & [\texttt{inv-child}, Z,Y]], \\
    [\texttt{un}, X,Y] &\vdash [[\texttt{sibling}, X,Z], \\ & [\texttt{child}, Z,Y]], \\
    [\texttt{inv-un}, X,Y] &\vdash [[\texttt{inv-child}, X,Z], \\ & [\texttt{sibling}, Z,Y]], \\
\end{split}
\end{align*}
\normalsize

\begin{figure*}[!ht]
     \centering
    \subfloat{{\includegraphics[width=0.48\textwidth]{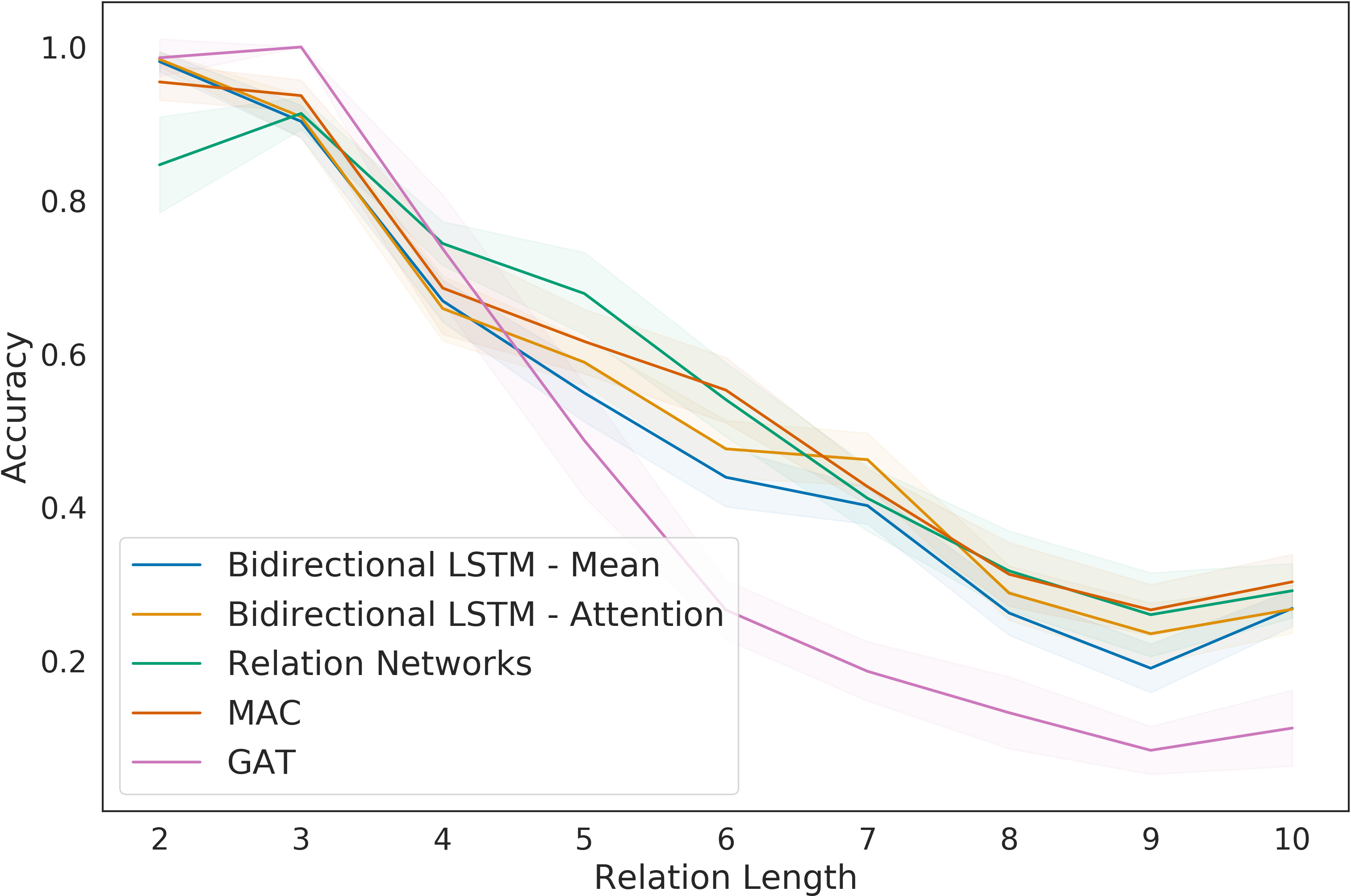} }}%
    \qquad
    \hspace{-20pt}
    \subfloat{{\includegraphics[width=0.48\textwidth]{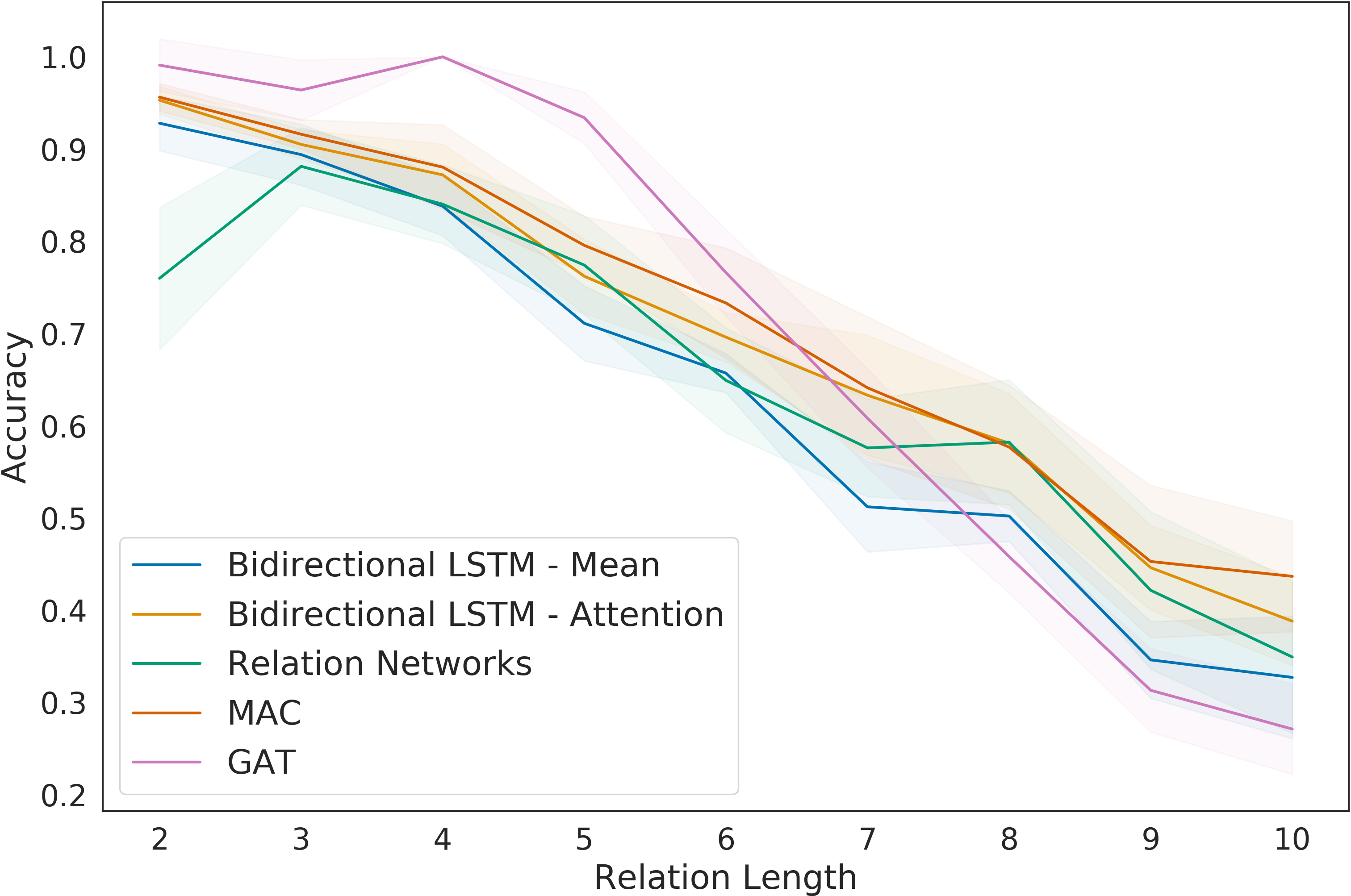} }}%
    \caption{Systematic Generalizability of different models on \texttt{CLUTRR-Gen} task (having 20\% less placeholders and without training and testing placeholder split), when {\bf Left:} trained with $k=2$ and $k=3$ and {\bf Right:} trained with $k=2,3$ and $4$}%
    \vspace{-10pt}
    \label{fig:gen_app_1}
\end{figure*}

In the CLUTRR Benchmark, the following kinship relations are used: \textit{son, father, husband, brother, grandson, grandfather, son-in-law, father-in-law, brother-in-law, uncle, nephew, daughter, mother, wife, sister, granddaughter, grandmother, daughter-in-law, mother-in-law, sister-in-law, aunt, niece}.

We used a small, tractable, and logically sound KB of rules as mentioned above. We carefully select this set of deterministic rules to avoid ambiguity in the resolution. We use gender-neutral predicates and resolve the gender of the predicate in the head $\mathcal{H}$ of a clause $\mathcal{C}$ by deducing the gender of the second constant. We have two types of predicates, \textit{vertical} predicates (parent-child relations) and  \textit{horizontal} predicates (sibling or significant other). We denote all the vertical predicates by its \textit{child-to-parent} relation and append the prefix \texttt{inv-} to the predicates for the corresponding \textit{parent-to-child} relation. For example, \texttt{grandfatherOf} is denoted by the gender-neutral predicate $[\texttt{inv-grand},X,Y]$, where the gender is determined by the gender of $Y$.

\subsection{Effect of placeholder size and split}
\label{app:easy_place_gen}

To analyze whether the language models fail to learn a robust mapping from natural language narratives to underlying logical facts, we re-run the generalization experiments with a reduced placeholder size (20\% of the full collected placeholders) \textit{and} we keep the same placeholders for both training and testing. We observe all language-based models are now competitive with respect to GAT on both training regimes $k=2,3$ and $k=2,3,4$. This shows the need for separating the placeholder split to effectively test systematic generalization because otherwise, current NLU systems tend to exploit the underlying language layer to arrive at the correct answer.

\subsection{More evaluations on Robust Reasoning}
\label{app:more_robust}

\begin{table*}[t!]
\resizebox{\textwidth}{!}{
% Please add the following required packages to your document preamble:
% \usepackage{multirow}
\begin{tabular}{@{}cccccccc|c@{}}
\toprule
\multicolumn{1}{l}{} & \multicolumn{1}{l}{Models} & \multicolumn{6}{c}{Unstructured models (no graph)} & \multicolumn{1}{l}{Structured model (with graph)} \\ \midrule
\multicolumn{1}{l|}{Training} & \multicolumn{1}{l|}{Testing} & BiLSTM - Attention & BiLSTM - Mean & RN & MAC & \multicolumn{1}{l}{BERT} & \multicolumn{1}{l|}{BERT-LSTM} & GAT \\ \midrule
Supporting & \multicolumn{1}{c|}{Clean} & 0.38 \tiny $\pm 0.04$ & 0.32 \tiny $\pm 0.04$ & 0.4 \tiny $\pm 0.09$ & 0.45 \tiny $\pm 0.03$ & 0.19 \tiny $\pm 0.06$ & 0.39 \tiny $\pm 0.06$ & \textbf{0.92} \tiny $\pm 0.17$ \\
\multicolumn{1}{l}{} & \multicolumn{1}{c|}{Supporting} & 0.67 \tiny $\pm 0.06$ & 0.66 \tiny $\pm 0.07$ & 0.68 \tiny $\pm 0.05$ & 0.65 \tiny $\pm 0.04$ & 0.32 \tiny $\pm 0.09$ & 0.57 \tiny $\pm 0.04$ & \textbf{0.98} \tiny $\pm 0.01$ \\
 & \multicolumn{1}{c|}{Irrelevant} & 0.44 \tiny $\pm 0.03$ & 0.39 \tiny $\pm 0.03$ & \textbf{0.51} \tiny $\pm 0.08$ & 0.46 \tiny $\pm 0.09$ & 0.2 \tiny $\pm 0.06$ & 0.36 \tiny $\pm 0.05$ & 0.5 \tiny $\pm 0.23$ \\
 & \multicolumn{1}{c|}{Disconnected} & 0.31 \tiny $\pm 0.21$ & 0.25 \tiny $\pm 0.16$ & 0.47 \tiny $\pm 0.08$ & 0.41 \tiny $\pm 0.06$ & 0.2 \tiny $\pm 0.08$ & 0.32 \tiny $\pm 0.04$ & \textbf{0.92} \tiny $\pm 0.05$ \\ \midrule
Irrelevant & \multicolumn{1}{c|}{Clean} & 0.57 \tiny $\pm 0.05$ & 0.56 \tiny $\pm 0.05$ & 0.46 \tiny $\pm 0.13$ & 0.67 \tiny $\pm 0.05$ & 0.24 \tiny $\pm 0.06$ & 0.46 \tiny $\pm 0.08$ & \textbf{0.92} \tiny $\pm 0.0$ \\
 & \multicolumn{1}{c|}{Supporting} & 0.38 \tiny $\pm 0.22$ & 0.31 \tiny $\pm 0.16$ & 0.61 \tiny $\pm 0.07$ & 0.61 \tiny $\pm 0.04$ & 0.27 \tiny $\pm 0.06$ & 0.46 \tiny $\pm 0.04$ & \textbf{0.77} \tiny $\pm 0.12$ \\
\multicolumn{1}{l}{} & \multicolumn{1}{c|}{Irrelevant} & 0.51 \tiny $\pm 0.06$ & 0.52 \tiny $\pm 0.06$ & 0.5 \tiny $\pm 0.04$ & 0.56 \tiny $\pm 0.04$ & 0.25 \tiny $\pm 0.06$ & 0.53 \tiny $\pm 0.06$ & \textbf{0.93} \tiny $\pm 0.01$ \\
 & \multicolumn{1}{c|}{Disconnected} & 0.44 \tiny $\pm 0.26$ & 0.54 \tiny $\pm 0.27$ & 0.55 \tiny $\pm 0.05$ & 0.61 \tiny $\pm 0.06$ & 0.26 \tiny $\pm 0.03$ & 0.45 \tiny $\pm 0.08$ & \textbf{0.85} \tiny $\pm 0.25$ \\ \midrule
\multirow{3}{*}{Disconnected} & \multicolumn{1}{c|}{Clean} & 0.45 \tiny $\pm 0.02$ & 0.47 \tiny $\pm 0.03$ & 0.53 \tiny $\pm 0.09$ & 0.5 \tiny $\pm 0.06$ & 0.22 \tiny $\pm 0.09$ & 0.44 \tiny $\pm 0.05$ & \textbf{0.75} \tiny $\pm 0.07$ \\
 & \multicolumn{1}{c|}{Supporting} & 0.47 \tiny $\pm 0.03$ & 0.46 \tiny $\pm 0.05$ & 0.54 \tiny $\pm 0.03$ & 0.58 \tiny $\pm 0.06$ & 0.22 \tiny $\pm 0.06$ & 0.38 \tiny $\pm 0.08$ & \textbf{0.78} \tiny $\pm 0.12$ \\
 & \multicolumn{1}{c|}{Irrelevant} & 0.47 \tiny $\pm 0.05$ & 0.48 \tiny $\pm 0.03$ & 0.52 \tiny $\pm 0.04$ & 0.51 \tiny $\pm 0.05$ & 0.17 \tiny $\pm 0.04$ & 0.38 \tiny $\pm 0.05$ & \textbf{0.56} \tiny $\pm 0.26$ \\
\multicolumn{1}{l}{} & \multicolumn{1}{c|}{Disconnected} & 0.57 \tiny $\pm 0.07$ & 0.57 \tiny $\pm 0.06$ & 0.45 \tiny $\pm 0.11$ & 0.4 \tiny $\pm 0.1$ & 0.17 \tiny $\pm 0.05$ & 0.47 \tiny $\pm 0.06$ & \textbf{0.96} \tiny $\pm 0.01$ \\ \midrule
\multicolumn{1}{l}{Average} & \multicolumn{1}{c|}{} & 0.47 \tiny $\pm 0.08$ & 0.46 \tiny $\pm 0.08$ & 0.52 \tiny $\pm 0.07$ & \textbf{0.53} \tiny $\pm 0.06$ & 0.23 \tiny $\pm 0.07$ & 0.43 \tiny $\pm 0.05$ & \textbf{0.82} \tiny $\pm 0.09$ \\ \bottomrule
\end{tabular}}
\caption{Testing the robustness of the various models when trained various types of noisy facts and evaluated on other noisy / clean facts. The types of noise facts (supporting, irrelevant and disconnected) are defined in Section 3.5 of the main paper.}
\label{tab:robust_appen}
\end{table*}

\begin{table*}[t!]
\resizebox{\textwidth}{!}{
% Please add the following required packages to your document preamble:
% \usepackage{multirow}
\begin{tabular}{@{}cccccccc|c@{}}
\toprule
\multicolumn{1}{l}{} & \multicolumn{1}{l}{Models} & \multicolumn{6}{c}{Unstructured models (no graph)} & \multicolumn{1}{l}{Structured model (with graph)} \\ \midrule
\multicolumn{1}{l|}{Training} & \multicolumn{1}{l|}{Testing} & BiLSTM - Attention & BiLSTM - Mean & RN & MAC & \multicolumn{1}{l}{BERT} & \multicolumn{1}{l|}{BERT-LSTM} & GAT \\ \midrule
Supporting & \multicolumn{1}{c|}{Clean} & 0.96 \tiny $\pm 0.01$ & \textbf{0.97} \tiny $\pm 0.01$ & 0.88 \tiny $\pm 0.05$ & 0.94 \tiny $\pm 0.02$ & 0.48 \tiny $\pm 0.08$ & 0.57 \tiny $\pm 0.08$ & 0.92 \tiny $\pm 0.17$ \\
\multicolumn{1}{l}{} & \multicolumn{1}{c|}{Supporting} & 0.96 \tiny $\pm 0.03$ & 0.96 \tiny $\pm 0.03$ & 0.97 \tiny $\pm 0.01$ & 0.97 \tiny $\pm 0.01$ & 0.75 \tiny $\pm 0.07$ & 0.88 \tiny $\pm 0.05$ & \textbf{0.98} \tiny $\pm 0.01$ \\
 & \multicolumn{1}{c|}{Irrelevant} & 0.92 \tiny $\pm 0.02$ & \textbf{0.93} \tiny $\pm 0.01$ & 0.9 \tiny $\pm 0.03$ & 0.91 \tiny $\pm 0.01$ & 0.56 \tiny $\pm 0.04$ & 0.54 \tiny $\pm 0.06$ & 0.5 \tiny $\pm 0.23$ \\
 & \multicolumn{1}{c|}{Disconnected} & 0.8 \tiny $\pm 0.04$ & 0.83 \tiny $\pm 0.04$ & 0.76 \tiny $\pm 0.08$ & 0.86 \tiny $\pm 0.04$ & 0.27 \tiny $\pm 0.06$ & 0.42 \tiny $\pm 0.08$ & \textbf{0.92} \tiny $\pm 0.05$ \\ \midrule
Irrelevant & \multicolumn{1}{c|}{Clean} & 0.63 \tiny $\pm 0.02$ & 0.61 \tiny $\pm 0.07$ & 0.85 \tiny $\pm 0.09$ & 0.8 \tiny $\pm 0.07$ & 0.53 \tiny $\pm 0.09$ & 0.44 \tiny $\pm 0.06$ & \textbf{0.92} \tiny $\pm 0.0$ \\
 & \multicolumn{1}{c|}{Supporting} & 0.66 \tiny $\pm 0.03$ & 0.64 \tiny $\pm 0.04$ & 0.69 \tiny $\pm 0.06$ & 0.76 \tiny $\pm 0.06$ & 0.42 \tiny $\pm 0.08$ & 0.43 \tiny $\pm 0.08$ & \textbf{0.77} \tiny $\pm 0.12$ \\
\multicolumn{1}{l}{} & \multicolumn{1}{c|}{Irrelevant} & 0.89 \tiny $\pm 0.04$ & 0.86 \tiny $\pm 0.1$ & 0.74 \tiny $\pm 0.11$ & 0.78 \tiny $\pm 0.06$ & 0.61 \tiny $\pm 0.1$ & 0.83 \tiny $\pm 0.06$ & \textbf{0.93} \tiny $\pm 0.01$ \\
 & \multicolumn{1}{c|}{Disconnected} & 0.64 \tiny $\pm 0.02$ & 0.62 \tiny $\pm 0.05$ & 0.72 \tiny $\pm 0.05$ & 0.73 \tiny $\pm 0.04$ & 0.41 \tiny $\pm 0.04$ & 0.61 \tiny $\pm 0.05$ & \textbf{0.85} \tiny $\pm 0.25$ \\ \midrule
\multirow{3}{*}{Disconnected} & \multicolumn{1}{c|}{Clean} & 0.9 \tiny $\pm 0.05$ & 0.82 \tiny $\pm 0.12$ & \textbf{0.94} \tiny $\pm 0.02$ & 0.93 \tiny $\pm 0.04$ & 0.68 \tiny $\pm 0.07$ & 0.64 \tiny $\pm 0.02$ & 0.75 \tiny $\pm 0.07$ \\
 & \multicolumn{1}{c|}{Supporting} & 0.87 \tiny $\pm 0.04$ & 0.82 \tiny $\pm 0.05$ & 0.85 \tiny $\pm 0.03$ & \textbf{0.88} \tiny $\pm 0.04$ & 0.54 \tiny $\pm 0.08$ & 0.5 \tiny $\pm 0.05$ & 0.78 \tiny $\pm 0.12$ \\
 & \multicolumn{1}{c|}{Irrelevant} & \textbf{0.87} \tiny $\pm 0.03$ & 0.85 \tiny $\pm 0.03$ & 0.83 \tiny $\pm 0.03$ & 0.87 \tiny $\pm 0.02$ & 0.59 \tiny $\pm 0.09$ & 0.58 \tiny $\pm 0.09$ & 0.56 \tiny $\pm 0.26$ \\
\multicolumn{1}{l}{} & \multicolumn{1}{c|}{Disconnected} & 0.91 \tiny $\pm 0.04$ & 0.91 \tiny $\pm 0.03$ & 0.8 \tiny $\pm 0.17$ & 0.71 \tiny $\pm 0.11$ & 0.49 \tiny $\pm 0.1$ & 0.79 \tiny $\pm 0.1$ & \textbf{0.96} \tiny $\pm 0.01$ \\ \midrule
\multicolumn{1}{l}{Average} & \multicolumn{1}{c|}{} & 0.83 \tiny $\pm 0.08$ & 0.82 \tiny $\pm 0.08$ & 0.83 \tiny $\pm 0.07$ & \textbf{0.84} \tiny $\pm 0.06$ & 0.58 \tiny $\pm 0.07$ & 0.60 \tiny $\pm 0.05$ & \textbf{0.82} \tiny $\pm 0.09$ \\ \bottomrule
\end{tabular}}
\caption{Testing the robustness on toy placeholders of the various models when trained various types of noisy facts and evaluated on other noisy / clean facts. The types of noise facts (supporting, irrelevant and disconnected) are defined in Section 3.5 of the main paper.}
\label{tab:robust_toy_appen}
\end{table*}

We performed several additional experiments to analyze the effect of different training regimes in the Robust Reasoning setup (Table \ref{tab:robust_appen}) of CLUTRR. Specifically, we want to analyze the effect on zero-shot generalization and robustness when training with different noisy data settings. We notice that the GAT model, having access to the true underlying graph of the puzzles, perform better across different testing scenarios when trained with the noisy data. As the \textit{Supporting facts} contains cycles, it is difficult for GAT to generalize for a dataset with cycles when it is trained on a dataset without cycles. However, when trained with cycles, GAT learns to attend to \textit{all} the paths leading to the correct answer. This effect is disastrous when GAT is tested on \textit{Irrelevant facts} which contains dangling paths as GAT still tries to attend to all the paths. Training on \textit{Irrelevant facts} proved to be most beneficial to GAT, as the model now perfectly attends to \textit{only relevant paths}.

Since \textit{Disconnected facts} contains disconnected paths, the message passing function of the graph is unable to forward any information from the disjoint cliques, thereby having superior testing scores throughout several scenarios.

\xhdr{Experiments on synthetic placeholders} In order to further understand the effect of language placeholders on robustness, we performed another set of experiments where we use bABI \cite{Weston2015-is} style simple placeholders (Table \ref{tab:robust_toy_appen}). We observe a marked increase in performance of all NLU models, where they significantly decrease the gap between their performance with that of GAT, even outperforming GAT on various settings. This shows the significance of using paraphrased placeholders in devising the complexity of the dataset.

%TODO: explain the data

\subsection{Comparison among different entity embedding policies}
\label{app:embeddings}

In Cloze style reading comprehension tasks, it is sometimes customary to choose UNK embeddings for entity placeholders. \cite{chen2016thorough} In our task, however, choosing UNK embeddings for entities is not feasible as the query involves two entities themselves. During preprocessing of our dataset, we convert the entity names into a Cloze-style setup where each entity is replaced by \textit{@entity-n} token. However, one has to be careful not to assign tokens in the same order for all the stories, which will lead to obvious overfitting since the models will learn to work around positional markers as shown in \citet{chen2016thorough}. Therefore, we randomize the Cloze-style entities themselves for each story. We experimented with three different policies of choosing the entity embeddings:

\begin{enumerate}
    \item \textit{Fixed Random Embeddings}: One simple and intuitive choice is to assign a random embedding to each entity and keep it fixed throughout the training. During our data-processing pipeline, we ensure that all the entity tokens are randomized using a pool of entity tokens, hence the chances of a model learning to exploit the positional markers are slim.
    \item \textit{Randomized Random Embeddings}: We can go one step further and randomize the random embeddings at \textit{each epoch}. This aggressive strategy does not let the model learn any positional markings at all, however it might hamper the learning ability of models as the entity representations are changing arbitrarily.
    \item \textit{Learned Random Embeddings}: Since our data pre-processing pipeline randomly assigns the entities on each story, we can as well learn a pool of $n$ entities, from which a subset is always used to replace the entities.
\end{enumerate}

\begin{figure*}[h]
     \centering
    \subfloat{{\includegraphics[width=0.48\textwidth]{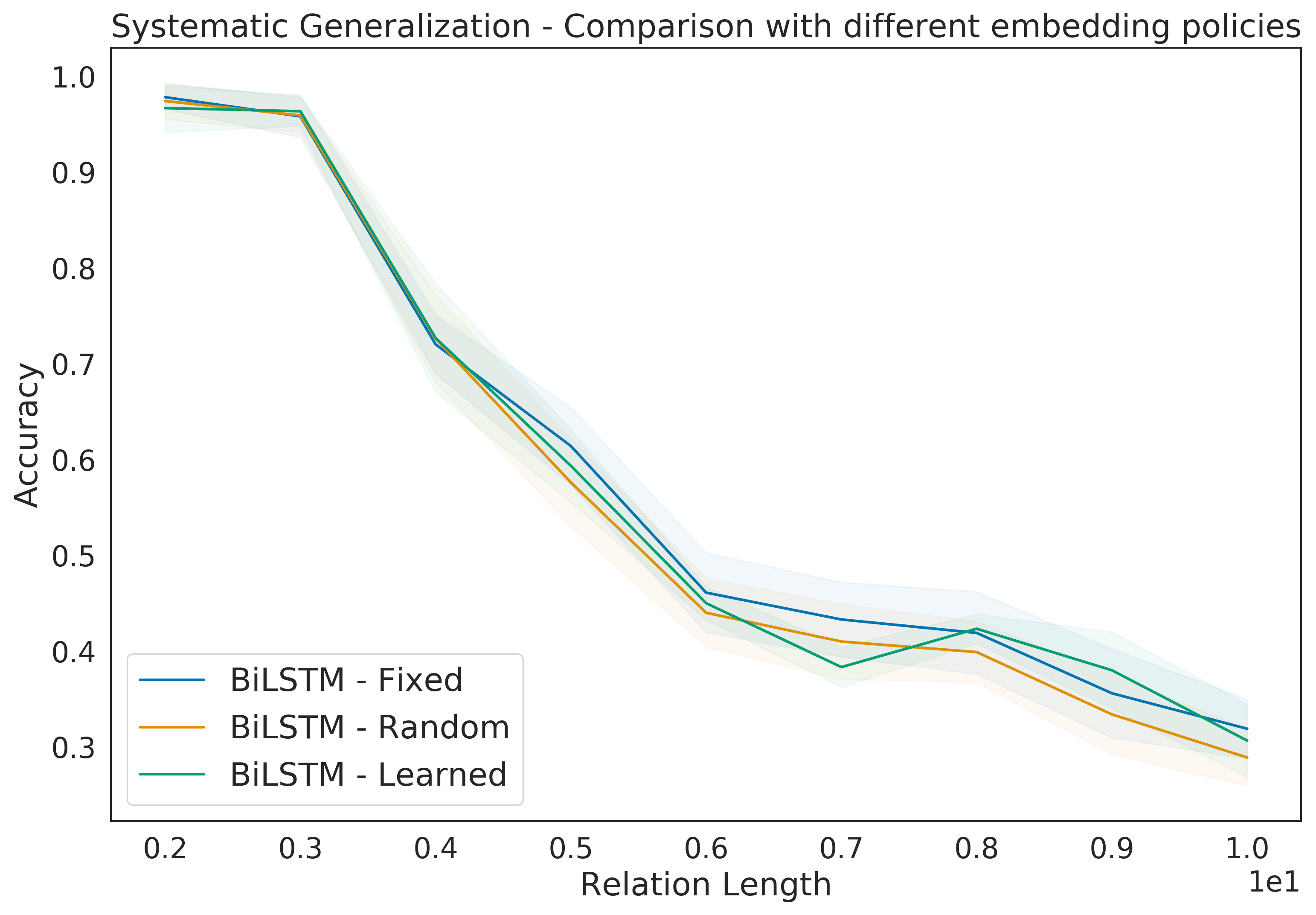} }}%
    \hspace{-20pt}
    \qquad
    \subfloat{{\includegraphics[width=0.48\textwidth]{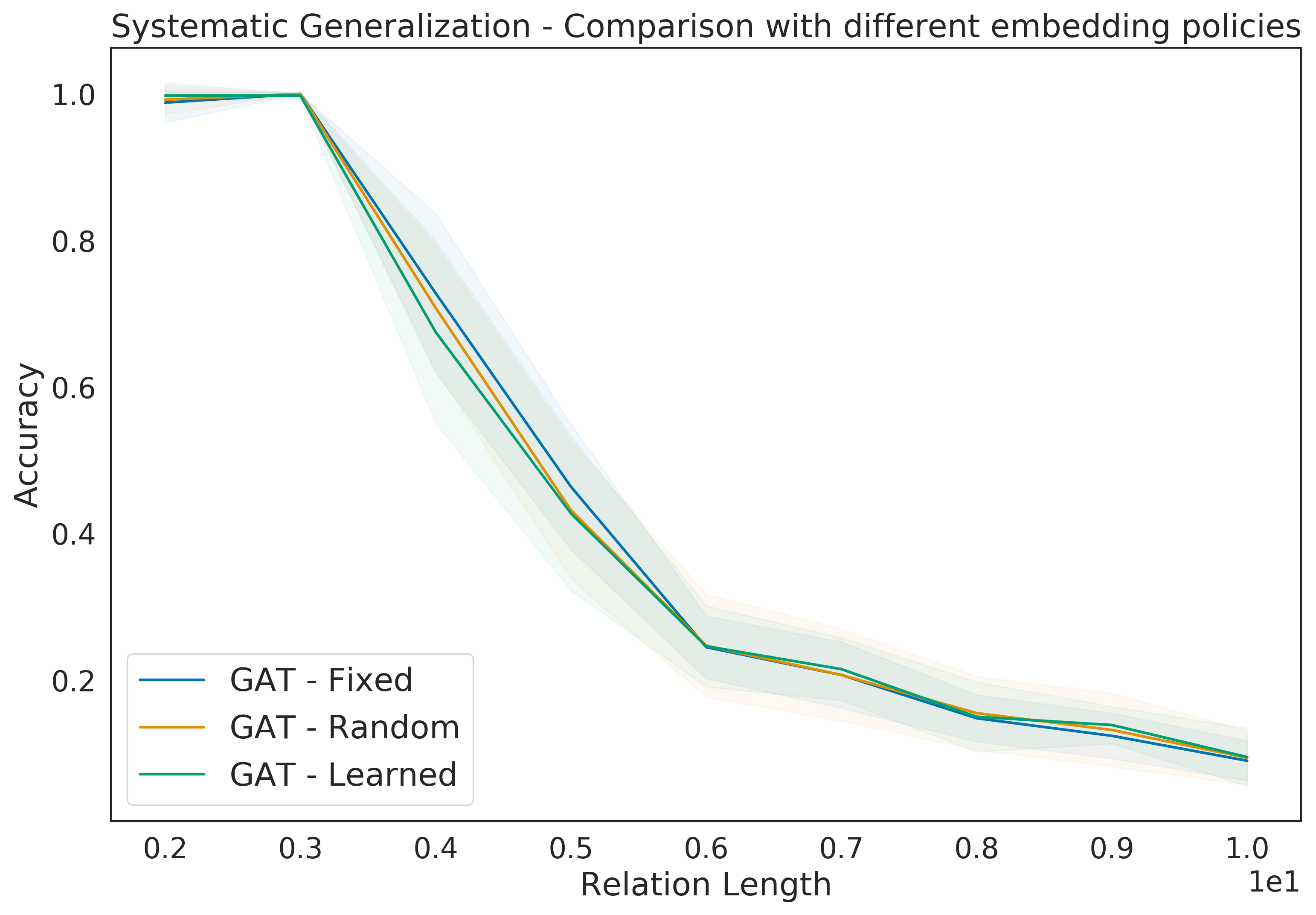} }}%
    \caption{Systematic Generalization comparison with different Embedding policies}%
    \vspace{-10pt}
    \label{fig:embedding_policy}
\end{figure*}

We chose to report all experiments with respect to fixed random embeddings. We compared different embedding policies with respect to the Systematic Generalization task. We show a comparison between the Bidirectional LSTM and GAT in Figure \ref{fig:embedding_policy}. We see that the fixed embedding policy has better Systematic Generalization score, although the advantage is minor compared to the other schemes. For GAT, the advantage is practically nil for the different schemes which shows that a Graph Neural Network performs inductive reasoning in the same manner irrespective of the initial node embedding representation.

\subsection{AMT Data collection process}
\label{app:amt_mturk}

We use ParlAI \cite{miller2017parlai} Mturk interface to collect paraphrases from the users. Specifically, given a set of facts, we ask the users to paraphrase the facts into a story. The users (\textit{turkers}) are free to construct any story they like as long as they mention all the entities and all the relations among them. We also provide the head $\mathcal{H}$ of the clause as an \textit{inferred} relation and specifically instruct the users to \textit{not} mention it in the paraphrased story. In order to evaluate the paraphrased stories, we ask the turkers to peer review a story paraphrased by a different turker. Since there are two tasks - paraphrasing a story and rating a story - we choose to pay 0.5\$ for each annotation. A sample task description in our MTurk interface is as follows:

\begin{quote}\small
    In this task, you will need to write a short, simple story based on a few facts. \textbf{It is crucial that the story mentions each of the given facts at least once.} The story does not need to be complicated! It just needs to be grammatical and mention the required facts.

    After writing the story, you will be asked to evaluate the quality of a generated story (based on a different set of facts). \textbf{It is crucial that you check whether the generated story mentions each of the required facts.}
    
    \textit{Example of good and bad stories: Good Example}
    
    \textbf{Facts to Mention}
    \begin{itemize}
        \item John is the father of Sylvia.
        \item Sylvia has a brother Patrick.
    \end{itemize}
    
    \textbf{Implied Fact}: John is the father of Patrick.
    
    \textbf{Written story}
    
    John is the proud father of the lovely Sylvia. Sylvia has a love-hate relationship with her brother Patrick.
    
    \textit{Bad Example}
    
    \textbf{Facts to Mention}
    
    \begin{itemize}
        \item Vincent is the son of Tim.
        \item Martha is the wife of Tim.
    \end{itemize}
    
    \textbf{Implied Fact} : Martha is Vincent's mother.
    
    \textbf{Written story}
    
    Vincent is married at Tim and his mother is Martha.
    
    \textit{The reason the above story is bad}:
    
    \begin{itemize}
        \item This story is bad because it is nonsense / ungrammatical.
        \item This story is bad because it does not mention the proper facts.
        \item This story is bad because it reveals the implied fact.
    \end{itemize}
\end{quote}

To ensure that the turkers are providing high-quality annotations without revealing the inferred fact, we also launch another task to ask the turkers to rate three annotations to be either good or bad which are provided by a set of \textit{different} turkers. We pay 0.2\$ for each HIT consisting of three reviews. This helped to remove logical and grammatical inconsistencies to a large extent. Based on the reviews, 79\% of the collected paraphrases passed the peer-review sanity check where all the reviewers agree on the quality. This subset of the placeholders is used in the benchmark. A sample of programmatically generated dataset for clause length of $k=2$ to $k=6$ is provided in the tables \ref{tab:puzzle_data_1} to \ref{tab:puzzle_data_5}.

\cut{
\subsection{Comparing the effect of adding linguistic diversity on synthetic data}
\label{app:amt_syn_comp}

\begin{figure*}[h]
     \centering
    \subfloat{{\includegraphics[width=0.48\textwidth]{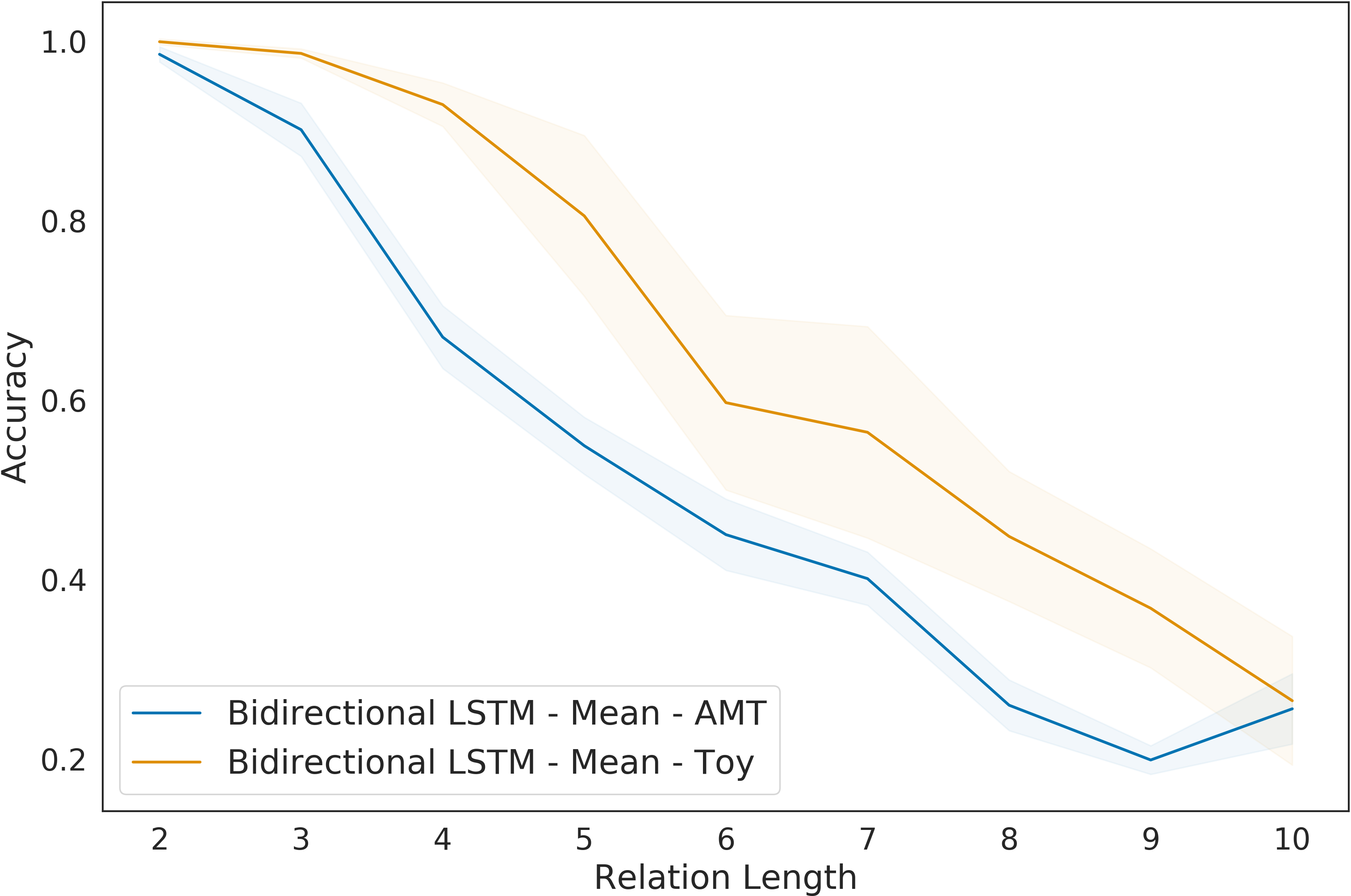} }}%
    \qquad
    \hspace{-20pt}
    \subfloat{{\includegraphics[width=0.48\textwidth]{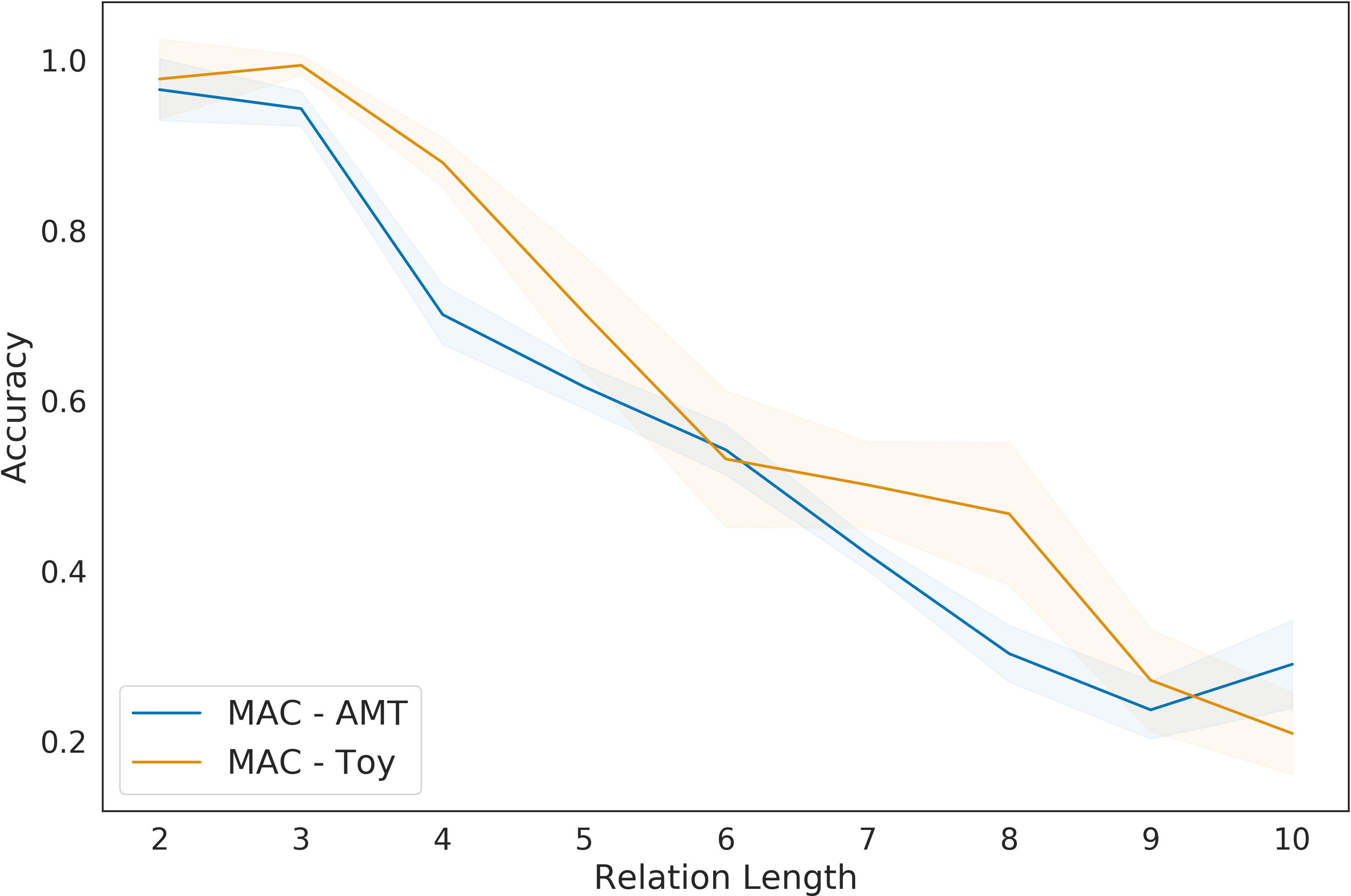} }}%
    \caption{Systematic Generalization performance on AMT paraphrased data}%
    \vspace{-10pt}
    \label{fig:amt_gen}
\end{figure*}

As we observe systematic generalization issues with the data generated from our benchmark suite, we want to analyze to what extent adding the linguistic diversity to the synthetic generated data helps. We performed the same experiments using the raw synthetic version of the dataset, \texttt{CLUTRR-Toy}. In terms of systematic generalization, we find on average the performance across model space is better on \texttt{CLUTRR-Toy}, which shows the inherent difficulty of compositional reasoning on natural language. Figure \ref{fig:amt_gen} compares the generalization performance between two such models, Bidirectional LSTM and MAC. 

\cut{
Surprisingly, when comparing performance on robust reasoning, most unstructured models display a boost in performance when trained with AMT paraphrased data compared to the synthetic data (Table \ref{tab:robust_toy}). This suggests that the linguistic diversity of the AMT data perhaps acts as a form of regularization, which affords increased robustness to the trained models.

\begin{table*}[t!]
\centering
\resizebox{\textwidth}{!}{
\begin{tabular}{|l|llll|lll|lll|lll|}
\hline
       &                                                  & \multicolumn{12}{c|}{Synthetic data}                                                                                                                                                                                                                                                              \\ \hline
       & \multicolumn{1}{l|}{Train}                       & \multicolumn{3}{c}{A}                                                  & \multicolumn{3}{c|}{B}                                                 & \multicolumn{3}{c|}{C}                                                 & \multicolumn{3}{c|}{D}                                                 \\ \cline{3-14} 
       & \multicolumn{1}{l|}{Test}                        & \multicolumn{1}{c}{B} & \multicolumn{1}{c}{C} & \multicolumn{1}{c|}{D} & \multicolumn{1}{c}{A} & \multicolumn{1}{c}{C} & \multicolumn{1}{c|}{D} & \multicolumn{1}{c}{A} & \multicolumn{1}{c}{B} & \multicolumn{1}{c|}{D} & \multicolumn{1}{c}{A} & \multicolumn{1}{c}{B} & \multicolumn{1}{c|}{C} \\ \hline
Model & \multicolumn{1}{l|}{BiLSTM - Attention}          & 0.693 \tiny \pm 0.01               & 0.717 \tiny \tiny \pm 0.04                & 0.706 \tiny \pm 0.03                  & 0.858 \tiny \pm 0.028                & 0.783 \tiny \pm 0.04                & 0.687 \tiny \pm 0.04                & 0.812 \tiny \pm 0.03                & 0.767 \tiny \pm 0.01                & 0.693 \tiny \pm 0.04                  & 0.897                 & 0.787           & 0.723                  \\
(no graph)      & \multicolumn{1}{l|}{BiLSTM - Mean}               & 0.708 \tiny  \pm 0.03               & 0.720 \tiny \pm 0.01                 & 0.676 \tiny  \pm 0.03                & 0.840 \tiny \pm 0.02                & 0.807 \tiny  \pm 0.02               & 0.702 \tiny \pm 0.06                 & 0.826 \tiny \pm 0.02                 & 0.787 \tiny \pm 0.03                & 0.657 \tiny \pm 0.04                  & 0.846                 & 0.786                 & 0.693                  \\
       & \multicolumn{1}{l|}{Relation Network}           & 0.698 \tiny \pm 0.02                 & 0.733  \tiny \pm 0.04               & 0.705 \tiny \pm 0.04                & 0.865     \tiny  \pm 0.07          & 0.882 \tiny  \pm 0.02               & 0.824 \tiny \pm 0.02                 & 0.866 \tiny \pm 0.05                & 0.849  \tiny \pm 0.02               & 0.805  \tiny \pm 0.04                & 0.853                 & 0.772                 & 0.725                  \\
       & \multicolumn{1}{l|}{MAC}                         & 0.770 \tiny \pm 0.04               & 0.773  \tiny \pm 0.02               & 0.747 \tiny \pm 0.05                 & 0.890 \tiny \pm 0.03                & 0.841 \tiny \pm 0.03                 & 0.720 \tiny \pm 0.04                 & 0.927             \tiny \pm 0.01    & 0.838 \tiny \pm 0.03                 & 0.742       \tiny \pm 0.03           & 0.893                 & 0.802                 & 0.713                  \\
       & \multicolumn{1}{l|}{Relational Recurrent Network} & 0.615 \tiny  \pm 0.03               & 0.650 \tiny \pm 0.02                & 0.561 \tiny \pm 0.04                 & 0.760 \tiny \pm 0.03               & 0.691 \tiny \pm 0.03                & 0.645       \tiny    \pm 0.05      & 0.738  \tiny \pm 0.03               & 0.688  \tiny \pm 0.02               & 0.608 \tiny \pm 0.04                 & 0.830                 & 0.609                 & 0.602                  \\ \hline
Graph Model       & \multicolumn{1}{l|}{GAT}                         & \textbf{0.811}                 & \textbf{0.803}                 & \textbf{0.855}                  & \textbf{0.990}                 & \textbf{0.952}                 & \textbf{0.969}                  & \textbf{0.999}                 & \textbf{0.984}                 & \textbf{0.949}                  & \textbf{0.992}                 & \textbf{0.833}                 & \textbf{0.910}                  \\ \hline
\end{tabular}}
\caption{Robustness testing on various scenarios of \texttt{CLUTRR-Toy}. Dataset instances are as follows: $A$ = \texttt{CLUTRR-Toy-Clean}, $B$ = \texttt{CLUTRR-Toy-Supporting}, $C$ = \texttt{CLUTRR-Toy-Irrelevant} and $D$ = \texttt{CLUTRR-Toy-Disconnected}}
\label{tab:robust_toy}
\end{table*}
}
}

\subsection{Human Evaluation}\label{app:human}

% Please add the following required packages to your document preamble:
% \usepackage{booktabs}
% \usepackage{multirow}
% \usepackage{graphicx}
\begin{table}[]
\centering
\resizebox{0.48\textwidth}{!}{%
\begin{tabular}{@{}llll@{}}
\toprule
\multirow{2}{*}{Relation Length} & \multicolumn{2}{l}{Human Performance} & \multirow{2}{*}{Reported Difficulty} \\
 & Time Limited & Unlimited Time &  \\ \midrule
2 & 0.848 & 1 & 1.488 +- 1.25 \\
3 & 0.773 & 1 & 2.41 +- 1.33 \\
4 & 0.477 & 1 & 3.81 +- 1.46 \\
5 & 0.424 & 1 & 3.78 +- 0.96 \\
6 & 0.406 & 1 & 4.46 +- 0.87 \\ \bottomrule
\end{tabular}%
}
\caption{Human performance accuracies on CLUTRR dataset. Humans are provided the Clean-Generalization version of the dataset, and we test on two scenarios: when a human is given limited time to solve the task, and when a human is given unlimited time to solve the task. Regardless of time, our evaluators provide a score of difficulty of individual puzzles.}
\label{tab:human_perf}
\end{table}

We performed a human evaluation study to analyze the difficulty of our proposed benchmark suite, which is provided in Table \ref{tab:human_perf}. We perform the evaluation in two scenarios: first a time-limited scenario where we ask AMT Turkers to solve the puzzle in a fixed time. Turkers were provided a maximum time of 30 mins, but they solved the puzzles in an average of 1 minute 23 seconds. Secondly, we use another set of expert evaluators who are given ample time to solve the tasks. Not surprisingly, if a human being is given ample time (experts took an average of 6 minutes per puzzle) and a pen and a paper to aid in the reasoning, they get all the relations correct. However, if an evaluator is short of time, they might miss important details on the relations and perform poorly. Thus, our tasks require \textit{active attention}.

In both cases, we asked Turkers and our expert human evaluators to rate the difficulty of a given task in a Likert scale of 1-5, where 1 corresponds to very easy and 5 corresponds to very hard perceived difficulty. This score increases as we increase the complexity of the task by increasing the relations, thereby suggesting that a human being perceives similar difficulty while solving for larger relation tasks. However, since a human being is a systematic learner, given enough time they can solve all puzzles with perfect accuracy. We set aside the task of testing noisy scenarios of CLUTRR to human evaluators as future work.

\section{Supplemental Material}
\label{sec:supplemental}
To promote reproducibility, we follow the guidelines proposed by the Machine Learning Reproducibility Checklist \footnote{\href{https://www.cs.mcgill.ca/~jpineau/ReproducibilityChecklist.pdf}{Machine Learning Reproducibility Checklist}} and release the following information regarding the experiments conducted by our benchmark suite.

\subsection{Details of datasets used}

A downloadable link to the datasets used can be found here \footnote{\href{https://drive.google.com/file/d/1SEq_e1IVCDDzsBIBhoUQ5pOVH5kxRoZF/view?usp=sharing}{Dataset link in Google Drive}}. Details of the individual datasets can be found in Table \ref{tab:public_data}. For all experiments, we use 10,000 training examples and a 100 testing example for each testing scenario. We split the training data 80-20 into a dev set randomly on each run.

\begin{table*}[]
\centering
\resizebox{\textwidth}{!}{%
\begin{tabular}{@{}lllll@{}}
\toprule
Dataset & Variant - Training & Variant - Testing & Training Clause length & Testing Clause length \\ \midrule
\texttt{data\_089907f8} & \multirow{2}{*}{Clean - Generalization} & \multirow{2}{*}{Clean - Generalization} & $(k=2,3)$ & \multirow{2}{*}{$(k=2,3, \ldots, 10)$} \\
\texttt{data\_db9b8f04} &  &  & $(k=2,3,4)$ &  \\ \midrule
\texttt{data\_7c5b0e70} & Clean & Clean, Supporting, Irrelevant, Disconnected & $(k=2,3)$ & $(k=2,3)$ \\ \midrule
\texttt{data\_06b8f2a1} & Supporting & Clean, Supporting, Irrelevant, Disconnected & $(k=2,3)$ & $(k=2,3)$ \\ \midrule
\texttt{data\_523348e6} & Irrelevant & Clean, Supporting, Irrelevant, Disconnected & $(k=2,3)$ & $(k=2,3)$ \\ \midrule
\texttt{data\_d83ecc3e} & Disconnected & Clean, Supporting, Irrelevant, Disconnected & $(k=2,3)$ & $(k=2,3)$ \\ \bottomrule
\end{tabular}%
}
\caption{Details of publicly released data}
\label{tab:public_data}
\end{table*}

\subsection{Details of Hyperparameters used}

For all models, the common hyperparameters used are: Embedding dimension: 100 (except BERT based models), Optimizer: Adam, Learning rate: 0.001, Number of epochs: 100, Number of runs: 10. Specific model-based hyperparameters are given as follows:

\begin{itemize}
    \item \textbf{Bidirectional LSTM}: LSTM hidden dimension: 100, \# layers: 2, Classifier MLP hidden dimension: 200
    \item \textbf{Relation Networks}: $f_{{\theta}_1}$ : 256, $f_{{\theta}_2}$: 64, $g_{\theta}$ : 64
    \item \textbf{MAC}: \# Iterations: 6, \texttt{shareQuestion}: True, Dropout - Memory, Read and Write: 0.2
    \item \textbf{Relational Recurrent Networks}: Memory slots: 2, Head size: 192, Number of heads: 4, Number of blocks : 1, forget bias : 1, input bias: 0, gate style: unit, key size: 64, \# Attention layers: 3, Dropout: 0
    \item \textbf{BERT}: Layers : 12, Fixed pretrained embeddings from \texttt{bert-base-uncased} using Pytorch HuggingFace BERT repository \footnote{\hyperlink{https://github.com/huggingface/pytorch-pretrained-BERT}{https://github.com/huggingface/pytorch-pretrained-BERT}}, Word dimension: 768, appended with a two-layer MLP for final prediction.
    \item \textbf{BERT-LSTM}: Same parameters as above, with a two-layer unidirectional LSTM encoder on top of BERT word embeddings.
    \item \textbf{GAT}: Node dimension: 100, Message dimension: 100, Edge dimension: 20, number of rounds: 3 
\end{itemize}

\begin{figure*}[h]
     \centering
    \includegraphics[width=\textwidth]{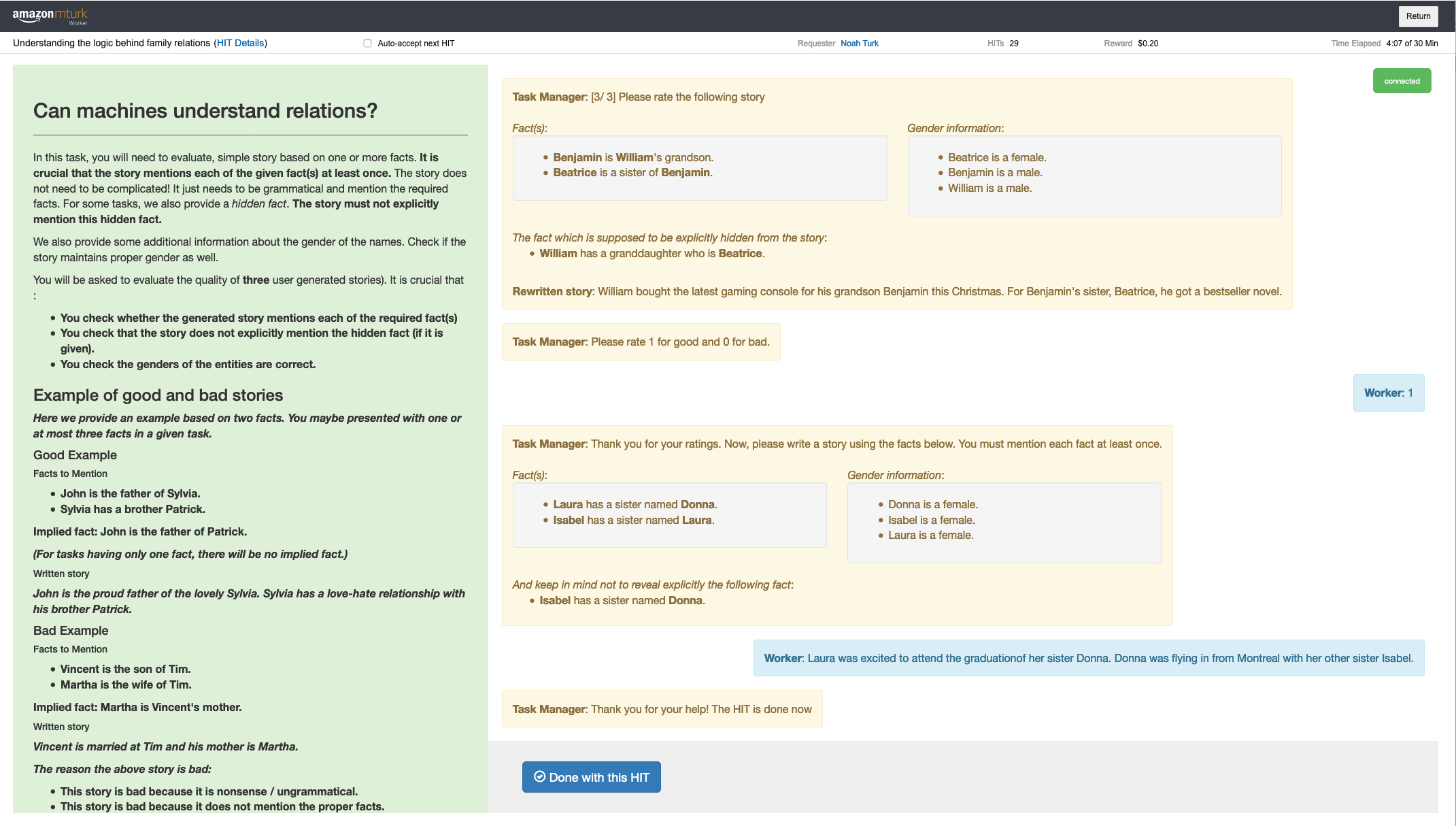}
    \caption{Amazon Mechanical Turker Interface built using ParlAI which was used to collect data as well as peer reviews.}%
    \vspace{-10pt}
    \label{fig:mturk}
\end{figure*}

\clearpage
% Please add the following required packages to your document preamble:
% \usepackage{booktabs}
% \usepackage{graphicx}
\begin{table*}[]
\centering
\caption{Snapshot of puzzles in the dataset for k=2}
\label{tab:puzzle_data_1}
\resizebox{\textwidth}{!}{%
\begin{tabular}{@{}p{7cm}lll@{}}
\toprule
\textbf{Puzzle} & \textbf{Question} & \textbf{Gender} & \textbf{Answer} \\ \midrule
\addlinespace[0.5em]
\noindent\parbox[c]{\hsize}{\textit{Charles}'s son \textit{Christopher} entered rehab for the ninth time at the age of thirty. \textit{Randolph} had a nephew called \textit{Christopher} who had n't seen for a number of years.} & Randolph is the \_\_\_\_\_ of Charles & \begin{tabular}[c]{@{}l@{}}Charles:male,\\ Christopher:male,\\ Randolph:male\end{tabular} & brother \\
\addlinespace[0.5em]
\noindent\parbox[c]{\hsize}{\textit{Randolph} and his sister \textit{Sharon} went to the park. \textit{Arthur} went to the baseball game with his son \textit{Randolph}}. & Sharon is the \_\_\_\_\_ of Arthur & \begin{tabular}[c]{@{}l@{}}Arthur:male,\\ Randolph:male,\\ Sharon:female\end{tabular} & daughter \\
\addlinespace[0.5em]
\noindent\parbox[c]{\hsize}{\textit{Frank} went to the park with his father, \textit{Brett}. \textit{Frank} called his brother \textit{Boyd} on the phone. He wanted to go out for some beers.} & Brett is the \_\_\_\_\_ of Boyd & \begin{tabular}[c]{@{}l@{}}Boyd:male,\\ Frank:male,\\ Brett:male\end{tabular} & father \\ \bottomrule
\end{tabular}%
}
\end{table*}

% Please add the following required packages to your document preamble:
% \usepackage{booktabs}
% \usepackage{graphicx}
\begin{table*}[]
\centering
\caption{Snapshot of puzzles in the dataset for k=3}
\label{tab:puzzle_data_2}
\resizebox{\textwidth}{!}{%
\begin{tabular}{@{}p{7cm}lll@{}}
\toprule
\textbf{Puzzle} & \textbf{Question} & \textbf{Gender} & \textbf{Answer} \\ \midrule
\textit{Roger} was playing baseball with his sons \textit{Sam} and \textit{Leon}. \textit{Sam} had to take a break though because he needed to call his sister \textit{Robin}. & Leon is the \_\_\_\_\_ of Robin & \begin{tabular}[c]{@{}l@{}}Robin:female,\\ Sam:male,\\ Roger:male,\\ Leon:male\end{tabular} & brother \\
\textit{Elvira} and her daughter \textit{Nancy} went shopping together last Monday and they bought new shoes for \textit{Elvira}'s kids. \textit{Pedro} and his sister \textit{Allison} went to the fair. \textit{Pedro}'s mother, \textit{Nancy}, was out with friends for the day. & Elvira is the \_\_\_\_\_ of Allison & \begin{tabular}[c]{@{}l@{}}Allison:female,\\ Pedro:male,\\ Nancy:female,\\ Elvira:female\end{tabular} & grandmother \\
\textit{Roger} met up with his sister \textit{Nancy} and her daughter \textit{Cynthia} at the mall to go shopping together. \textit{Cynthia}'s brother \textit{Pedro} was going to be the star in the new show. & Pedro is the \_\_\_\_\_ of Roger & \begin{tabular}[c]{@{}l@{}}Roger:male,\\ Nancy:female,\\ Cynthia:female,\\ Pedro:male\end{tabular} & nephew \\ \bottomrule
\end{tabular}%
}
\end{table*}

\clearpage
% Please add the following required packages to your document preamble:
% \usepackage{booktabs}
% \usepackage{graphicx}
\begin{table*}[]
\centering
\caption{Snapshot of puzzles in the dataset for k=4}
\label{tab:puzzle_data_3}
\resizebox{\textwidth}{!}{%
\begin{tabular}{@{}p{7cm}lll@{}}
\toprule
\textbf{Puzzle} & \textbf{Question} & \textbf{Gender} & \textbf{Answer} \\ \midrule
\textit{Celina} has been visiting her sister, \textit{Fran} all week. \textit{Fran} is also the daughter of \textit{Bethany}. \textit{Ronald} loves visiting his aunt \textit{Bethany} over the weekends. \textit{Samuel}'s son \textit{Ronald} entered rehab for the ninth time at the age of thirty. & Celina is the \_\_\_\_\_ of Samuel & \begin{tabular}[c]{@{}l@{}}Samuel:male,\\ Ronald:male,\\ Bethany:female,\\ Fran:female,\\ Celina:female\end{tabular} & niece \\
\textit{Celina} adores her daughter \textit{Bethany}. \textit{Bethany} loves her very much, too. \textit{Jackie} called her mother \textit{Bethany} to let her know she will be back home soon. \textit{Thomas} was helping his daughter \textit{Fran} with her homework at home. Afterwards, \textit{Fran} and her sister \textit{Jackie} played Xbox together. & Celina is the \_\_\_\_\_ of Thomas & \begin{tabular}[c]{@{}l@{}}Thomas:male,\\ Fran:female,\\ Jackie:female,\\ Bethany:female,\\ Celina:female\end{tabular} & daughter \\
\textit{Raquel} is \textit{Samuel} 'daughter and they go shopping at least twice a week together. \textit{Kenneth} and her mom, \textit{Theresa}, had a big fight. \textit{Theresa}'s son, \textit{Ronald}, refused to get involved. \textit{Ronald} was having an argument with her sister, \textit{Raquel}. & Samuel is the \_\_\_\_\_ of Kenneth & \begin{tabular}[c]{@{}l@{}}Kenneth:male,\\ Theresa:female,\\ Ronald:male,\\ Raquel:female,\\ Samuel:male\end{tabular} & father \\ \bottomrule
\end{tabular}%
}
\end{table*}

\clearpage
% Please add the following required packages to your document preamble:
% \usepackage{booktabs}
% \usepackage{graphicx}
\begin{table*}[]
\centering
\caption{Snapshot of puzzles in the dataset for k=5}
\label{tab:puzzle_data_4}
\resizebox{\textwidth}{!}{%
\begin{tabular}{@{}p{7cm}lll@{}}
\toprule
\textbf{Puzzle} & \textbf{Question} & \textbf{Gender} & \textbf{Answer} \\ \midrule
\textit{Steven}'s son is \textit{Bradford}. \textit{Bradford} and his father always go fishing together on Sundays and have a great time together. \textit{Diane} is taking her brother \textit{Brad} out for a late dinner. \textit{Kristin}, \textit{Brad}'s mother, is home with a cold. \textit{Diane}'s father \textit{Elmer}, and his brother \textit{Steven}, all got into the rental car to start the long cross-country roadtrip they had been planning. & Bradford is the \_\_\_\_\_ of Kristin & \begin{tabular}[c]{@{}l@{}}Kristin:female,\\ Brad:male,\\ Diane:female,\\ Elmer:male,\\ Steven:male,\\ Bradford:male\end{tabular} & nephew \\
\textit{Elmer} went on a roadtrip with his youngest child, \textit{Brad}. \textit{Lena} and her sister \textit{Diane} are going to a restaurant for lunch. \textit{Lena}'s brother \textit{Brad} is going to meet them there with his father \textit{Elmer} \textit{Brad} ca n't stand his unfriendly aunt \textit{Lizzie}. & Lizzie is the \_\_\_\_\_ of Diane & \begin{tabular}[c]{@{}l@{}}Diane:female,\\ Lena:female,\\ Brad:male,\\ Elmer:male,\\ Lizzie:female\end{tabular} & aunt \\
\textit{Ira} took his niece \textit{April} fishing Saturday. They caught a couple small fish. \textit{Ronald} was enjoying spending time with his parents, \textit{Damion} and \textit{Claudine}. \textit{Damion}'s other son, \textit{Dennis}, wanted to come visit too. \textit{Dennis} often goes out for lunch with his sister, \textit{April}. & Ira is the \_\_\_\_\_ of Claudine & \begin{tabular}[c]{@{}l@{}}Claudine:female,\\ Ronald:male,\\ Damion:male,\\ Dennis:male,\\ April:female,\\ Ira:male\end{tabular} & brother \\ \bottomrule
\end{tabular}%
}
\end{table*}

\clearpage
% Please add the following required packages to your document preamble:
% \usepackage{booktabs}
% \usepackage{graphicx}
\begin{table*}[]
\centering
\caption{Snapshot of puzzles in the dataset for k=6}
\label{tab:puzzle_data_5}
\resizebox{\textwidth}{!}{%
\begin{tabular}{@{}p{7cm}lll@{}}
\toprule
\textbf{Puzzle} & \textbf{Question} & \textbf{Gender} & \textbf{Answer} \\ \midrule
\textit{Mario} wanted to get a good gift for his sister, \textit{Marianne}. \textit{Jean} and her sister \textit{Darlene} were going to a party held by \textit{Jean}'s mom, \textit{Marianne}. \textit{Darlene} invited her brother \textit{Roy} to come, too, but he was too busy. \textit{Teri} and her father, \textit{Mario}, had an argument over the weekend. However, they made up by Monday. \textit{Agnes} wants to make a special meal for her daughter \textit{Teri}'s birthday. & Roy is the \_\_\_\_\_ of Agnes & \begin{tabular}[c]{@{}l@{}}Agnes:female,\\ Teri:female,\\ Mario:male,\\ Marianne:female,\\ Jean:female,\\ Darlene:female,\\ Roy:male\end{tabular} & nephew \\
\textit{Robert}'s aunt, \textit{Marianne}, asked \textit{Robert} to mow the lawn for her. \textit{Robert} said he could n't because he had a bad back. \textit{William}'s parents, \textit{Brian} and \textit{Marianne}, threw him a surprise party for his birthday. \textit{Brian}'s daughter \textit{Jean} made a mental note to be out of town for her birthday! \textit{Agnes}'s biggest accomplishment is raising her son \textit{Robert}. \textit{Jean} is looking for a good gift for her sister \textit{Darlene}. & Darlene is the \_\_\_\_\_ of Agnes & \begin{tabular}[c]{@{}l@{}}Agnes:female,\\ Robert:male,\\ Marianne:female,\\ William:male,\\ Brian:male,\\ Jean:female,\\ Darlene:female\end{tabular} & niece \\
\textit{Sharon} and her brother \textit{Mario} went shopping. \textit{Teri}, \textit{Mario}'s daughter, came too. \textit{Agnes}, \textit{Annie}'s mother, is unhappy with \textit{Robert}. She feels her son is cruel to \textit{Annie}'s sister \textit{Teri}, and she wants \textit{Robert} to be nicer. \textit{Robert}'s sister, \textit{Nicole}, participated in the dance contest. & Nicole is the \_\_\_\_\_ of Sharon & \begin{tabular}[c]{@{}l@{}}Sharon:female,\\ Mario:male,\\ Teri:female,\\ Annie:female,\\ Agnes:female,\\ Robert:male,\\ Nicole:female\end{tabular} & niece \\ \bottomrule
\end{tabular}%
}
\end{table*}

\end{document}